\documentclass[journal]{IEEEtran}

\usepackage{graphics} 
\usepackage{epsfig} 
\usepackage{balance}
\usepackage{amsmath} 
\usepackage{amssymb}  

\usepackage[para,online,flushleft]{threeparttable}
\usepackage[english]{babel}
\usepackage[T1]{fontenc}
\usepackage[utf8]{inputenc}
\addto\captionsenglish{\renewcommand{\figurename}{Fig.}}

\usepackage{booktabs} 
\usepackage{verbatim}
\usepackage{cite}
\usepackage[caption=false,font=footnotesize]{subfig}
\usepackage{multirow}
\usepackage{tabularx}
\usepackage{tablefootnote}
\usepackage{footnote}
\usepackage{dblfloatfix}
\usepackage{amsmath}
\usepackage[normalem]{ulem} 
\usepackage{xparse,soul} 
\soulregister\cite7 
\soulregister\citep7 
\soulregister\citet7 
\soulregister\ref7 
\soulregister\pageref7 
\renewcommand{\hl}{} 
\usepackage{xcolor}
\usepackage{lineno}
\usepackage[normalem]{ulem}

\usepackage[explicit]{titlesec}

\titlespacing\section{0pt}{5pt plus 2pt minus 2pt}{0pt plus 0pt minus 0pt}
\titlespacing\subsection{0pt}{5pt plus 2pt minus 2pt}{0pt plus 0pt minus 0pt}
\titlespacing\subsubsection{0pt}{0pt plus 2pt minus 2pt}{0pt plus 0pt minus 0pt}

\usepackage[linesnumbered, ruled]{algorithm2e}
\SetKwRepeat{Do}{do}{while}%

\usepackage{tikz}
\usepackage{textcomp}
\usepackage{lipsum}
\usepackage[bookmarks=false]{hyperref}

\usepackage[linesnumbered, ruled]{algorithm2e}
\SetKwRepeat{Do}{do}{while}
\let\oldnl\nl
\newcommand{\nonl}{\renewcommand{\nl}{\let\nl\oldnl}}

\ifCLASSINFOpdf
\else
\fi
\begin{document}
%
\title{Analytical Design and Development of a Modular \\ and Intuitive  Framework for Robotizing and \\ Enhancing  the  Existing  Endoscopic Procedures}

%

\author{Mohammad Rafiee Javazm$^{1}$, Yash Kulkarni$^{1}$, Jiaqi Xue$^{1}$, 
\\ Naruhiko Ikoma$^{2}$, \IEEEmembership{~Member,~IEEE}, and Farshid Alambeigi$^{1}$,\IEEEmembership{~Member,~IEEE}
		\thanks{	*Research reported in this publication was supported by the National Cancer Institute of the National Institutes of Health under Award Number R21CA280747.}
			\thanks{$^{1}$M. R. Javazm, Y.~Kulkarni,  J.~Xue, and F.~Alambeigi are with the Walker Department of Mechanical Engineering and the Texas Robotics at the University of Texas at Austin, Austin, TX, 78712, USA. Email: {\tt\footnotesize \{mohammad.rafiee, kulkarni.yash08, jiaqixue\}@utexas.edu,  farshid.alambeigi@austin.utexas.edu}.}
 	\thanks{$^{2}$N. Ikoma is with the Department of Surgical Oncology, Division of Surgery, The University of Texas MD Anderson Cancer Center, Houston, TX, 77030, USA. Email:
{\tt\footnotesize  nikoma@mdanderson.org}.}}
\maketitle

\begin{abstract}   
Despite the widespread adoption of endoscopic devices for several cancer screening procedures, manual control of these devices still remains challenging for clinicians, leading to several critical issues such as increased workload, fatigue, and distractions. 
To address these issues, in this paper,  we introduce the design and development of an intuitive, modular, and easily installable mechatronic framework. This framework includes (i) a novel nested collet-chuck gripping mechanism that can readily be integrated and assembled with the existing endoscopic devices and control their bending degrees-of-freedom (DoFs); (ii) a feeder mechanism that can control the insertion/retraction DoF of a colonoscope, and (iii) a complementary and intuitive user interface that enables simultaneous control of all DoFs during the procedure. To analyze the design of the proposed mechanisms, we also introduce a mathematical modeling approach and a design space for optimal selection of the parameters involved in the design of gripping and feeder mechanisms. Our simulation and experimental studies thoroughly demonstrate the performance of the proposed mathematical modeling and robotic framework.      
\end{abstract}

\begin{IEEEkeywords}
	Modular Mechatronics System, Steerable Colonoscope/Endoscope, Colorectal Cancer Screening.
\end{IEEEkeywords}

\IEEEpeerreviewmaketitle

\section{INTRODUCTION}\label{secintro}
\IEEEPARstart{S}{ince} their invention in late 1800's, endoscopic devices have evolved into  versatile and multi-functional instruments, surpassing their original role in  examining complex anatomies (e.g., urethra and gastrointestinal (GI) tract) to  a platform used for diverse medical interventions such as biopsy and  polypectomy~\cite{Williams2009}. For example, colonoscopy, as an endoscopic procedure, is the gold standard  of colorectal cancer (CRC) screening, which is the prominent cause of cancer-related deaths worldwide~\cite{sung2021global}. 
Despite their tremendous benefits, literature states that colonoscopic procedures suffer from an early detection miss rate of approximately 30\% for CRC polyps\cite{zhao2019magnitude}. This high detection miss rate can be attributed to non-optimal mechanical design, non-intuitive control interface, slow learning curve,  and limitations of the used optical sensing system (e.g., occlusion and blur) in this endoscopic device~\cite{azer2019challenges}.
While researchers and companies have made significant strides in enhancing endoscopic technologies with advanced optical sensors/cameras~\cite{Alian2022CurrentED,Gerald2022ASR,Kim2023ASA}, there has been a noticeable gap in addressing the steerability issues and intuitiveness of controlling these devices~\cite{Williams2009}. Surprisingly, after more than a  century, the overall mechanical structure and control interface of these devices have not gone through significant changes since their invention.

\begin{figure}[t]
    \centering
    \includegraphics[width=1\columnwidth]{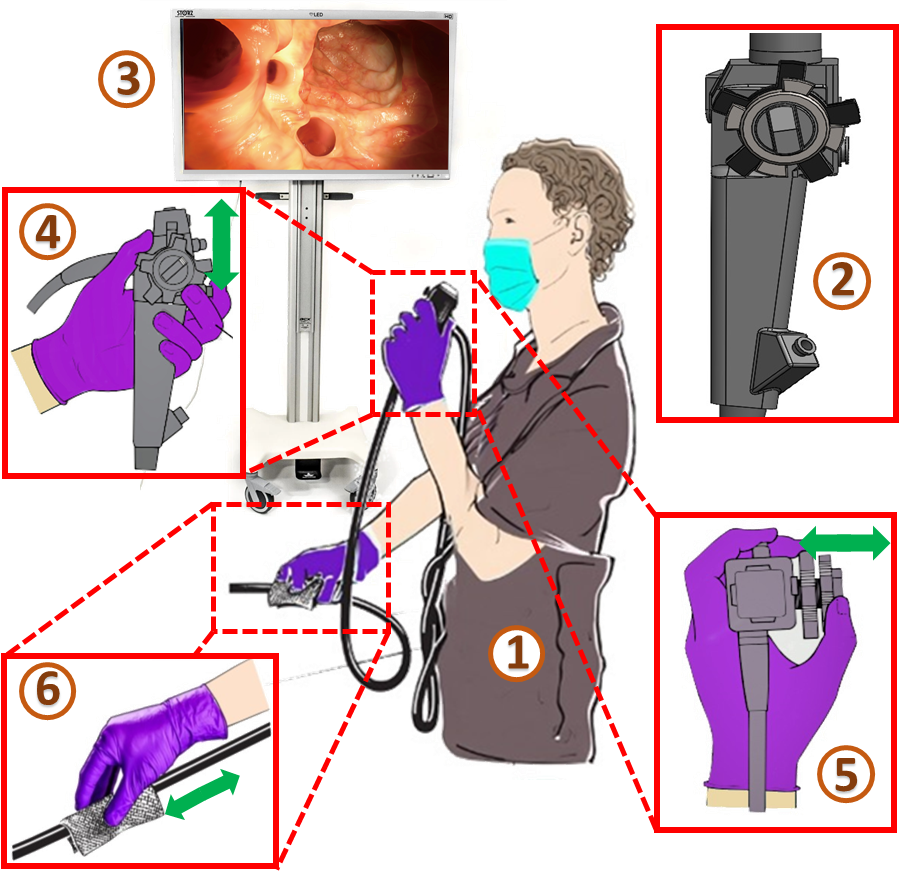}
     \vspace*{-7mm}
     \caption{Illustration of conventional colonoscopy procedure for CRC diagnosis: \raisebox{.5pt}{\textcircled{\raisebox{-.9pt} {1}}} - A surgeon holding a colonoscope
     , \raisebox{.5pt}{\textcircled{\raisebox{-.9pt} {2}}} - Close-up view of the handheld knob of colonoscope
     , \raisebox{.5pt}{\textcircled{\raisebox{-.9pt} {3}}} - A monitor for screening, \raisebox{.5pt}{\textcircled{\raisebox{-.9pt} {4}}} - demonstrating the procedure of holding and manually navigating the colonoscope 
     by a skillful surgeon for bending the end effector in the Upward/Downward direction, \raisebox{.5pt}{\textcircled{\raisebox{-.9pt} {5}}} - Left/Right direction, and \raisebox{.5pt}{\textcircled{\raisebox{-.9pt} {6}}} - Insertion/Retraction of colonoscope.}
    \label{fig:Procedure}
\end{figure}
As shown in Fig.\ref{fig:Procedure},
to perform an endoscopic procedure, a clinician uses the \textit{control handle} (CH) of this device and pulling/pushing of tube to actuate three degrees of freedom (DoFs). 
Since the required DoFs to perform the procedure is more than  clinicians' hands (i.e., 2 versus 3), a manual locking lever has also been considered to first lock one of the controlled DoFs and then control the other one when needed.
This sequential locking of the DOFs can make  control of these flexible  instruments --  that are steered inside a deformable  anatomy (e.g., a colon and esophagus) -- very challenging and non-intuitive. This problem becomes more critical, when a clinician decides to hold the position of the endoscope and perform a specific procedure such as a biopsy or polypectomy. 
The cumbersome and non-intuitive handling of these flexible devices directly affects (i) physical and mental fatigue of clinicians and thereby the quality of colonoscopic procedures
,  (ii) extended duration of the procedure, and (iii) patient's healthcare. These limitations, therefore, demand design and development of ergonomic and user-centric colonoscopic devices with easy-to-use control interfaces. 

\indent To address the aforementioned challenges of existing endoscopes, some researchers have opted to develop novel robotic endoscopes (e.g., \cite{Kume2015DevelopmentOA,Takamatsu2023RoboticEW}), whereas others tried to utilize the existing endoscopic device and improve their steerability by motorizing their CH (e.g.,\cite{Reilink2010EndoscopicCC,Vrielink2018IntuitiveGO,Lee2020easyEndoRE,Basha2023AGS}).
For example, Kume et al. \cite{Kume2015DevelopmentOA} developed a four DoF robotic system for colonoscopic procedures that provides haptic feedback.
Takamatsu et al. \cite{Takamatsu2023RoboticEW} also proposed a robotic colonoscope featuring a double-balloon and double-bend tube to enhance steerability during the insertion and reduce patient discomfort. 
While robotic endoscopic systems have shown potential, they often come with high complexity and cost, lack modularity for easy integration with existing devices, and face challenges in real-world clinical workflows. 

On the other hand,  Reilink et al. \cite{Reilink2010EndoscopicCC} presented relatively large actuation system based on toothed belts to control an endoscopic camera using head movements, aiming to eliminate the need for an assistant during surgery. \cite{Iwasa2018ANR} has also presented the prototype of a flexible robotic endoscope with single-hand control. The functionality of this device has been evaluated in endoscopic submucosal dissection of porcine stomachs through a tele-operation procedure. 
Vrielink et al. \cite{Vrielink2018IntuitiveGO} developed an intuitive gaze-control to automate endoscopy procedure, allowing easier navigation and manipulation. Basha et al. \cite{Basha2023AGS} proposed a generic scope actuation system to enhance the maneuverability of flexible endoscopes. Further, Lee et al. \cite{Lee2020easyEndoRE} introduced a mechanism called easyEndo to automate and enable single-hand control of existing commercialized endoscopes by integrating with its CH. Despite the success of the performed experiments, the complexity of the proposed mechanism makes its assembly, fabrication, and integration with the existing surgical workflow very cumbersome. 

\begin{figure}[h!]
   \centering \includegraphics[width=1\columnwidth]{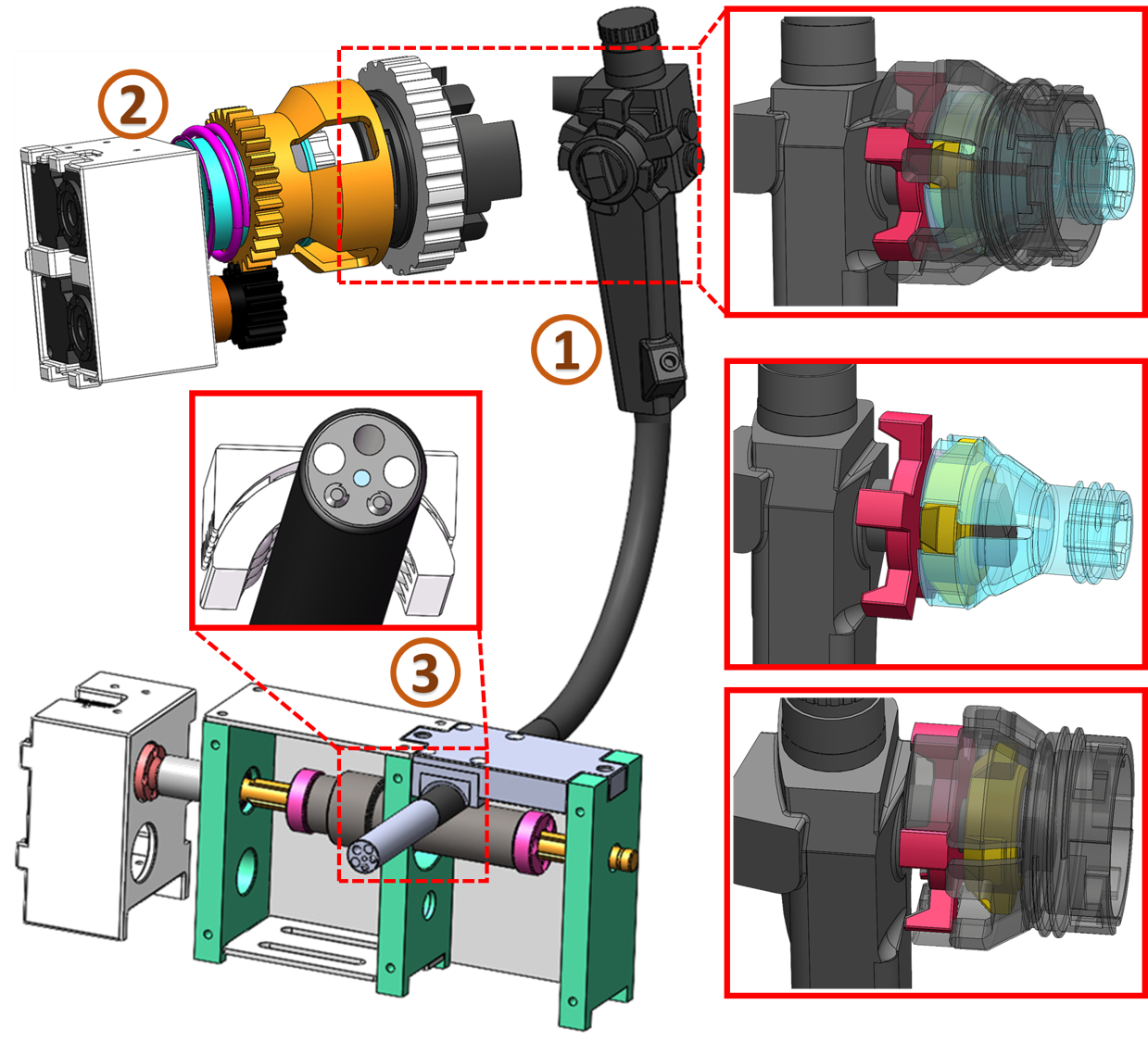}
    \vspace*{-8mm}
    \caption{Overview of our proposed design, including \raisebox{.5pt}{\textcircled{\raisebox{-.9pt} {1}}} - A  commercialized colonoscope, \raisebox{.5pt}{\textcircled{\raisebox{-.9pt} {2}}} - The proposed CH Gripping Mechanism to handle two bending DOFs of colonoscope, and \raisebox{.5pt}{\textcircled{\raisebox{-.9pt} {3}}} - The proposed Feeder Mechanism to provide insertion/retraction of colonoscope. Figure also shows a close-up of the collet mechanism, which securely holds the CH as well as a zoomed view of the sliding mechanism within the feeder, intended to secure the scope while allowing for smooth and free sliding.}
    \label{fig:EasyC-Basic}
\end{figure}

\indent To address the above-mentioned challenges in the non-intuitive interface of manual control and the lack of ergonomic support of current colonoscopes, in this study and as our main \textit{contributions}, we propose  a novel mechatronics framework that can seamlessly be integrated with the CH of the existing commercialized  endoscopic devices  regardless of their  size differences and manufacturers. As shown in Fig.\ref{fig:EasyC-Basic}, this unique framework includes (i) a novel nested collet chuck gripping mechanism that can readily be integrated and assembled with the CH of a generic endoscope made by different manufacturers  (e.g., Olympus and Pentax) to control its bending DoFs; (ii) a feeder mechanism that can control the insertion/retraction DoF of these endoscopes, and (iii) a complementary and intuitive user interface that enables simultaneous  control of all DoFs during the procedure.
\begin{figure*}[h!]
   \centering
    \includegraphics[width=2\columnwidth]{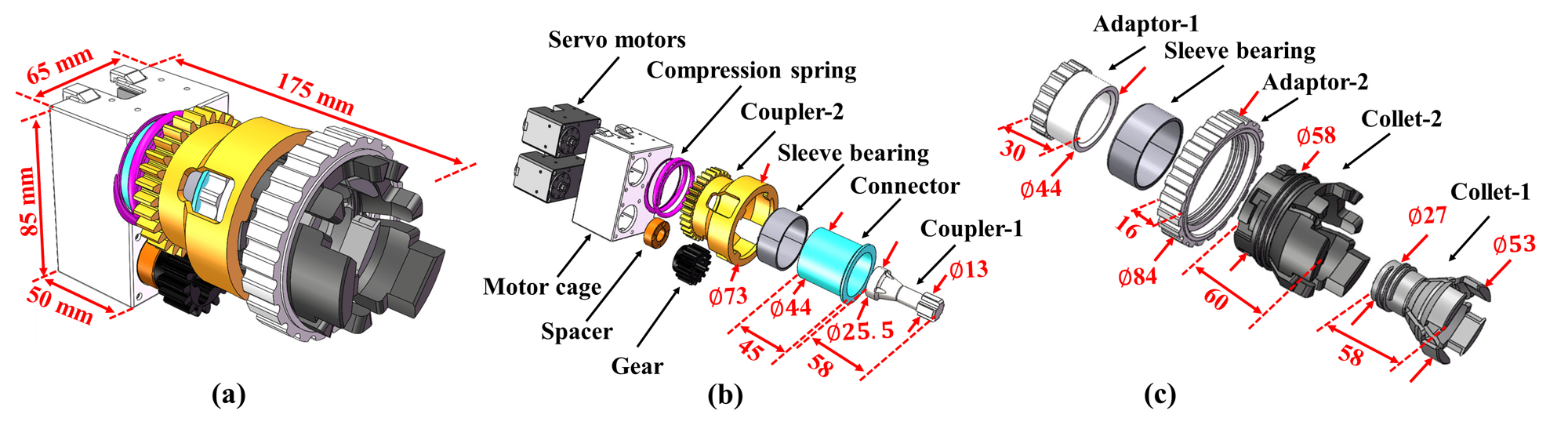}
     \vspace*{-4mm}
     \caption{CH Gripping Mechanism: (a) Assembly view, (b) Exploded view of power transmission, and (c) Exploded view of grasping components. All dimensions are in millimeter (mm).} 
    \label{fig:BendingMechanism}
\end{figure*}

\begin{figure}[t]
   \centering \includegraphics[width=1\columnwidth]{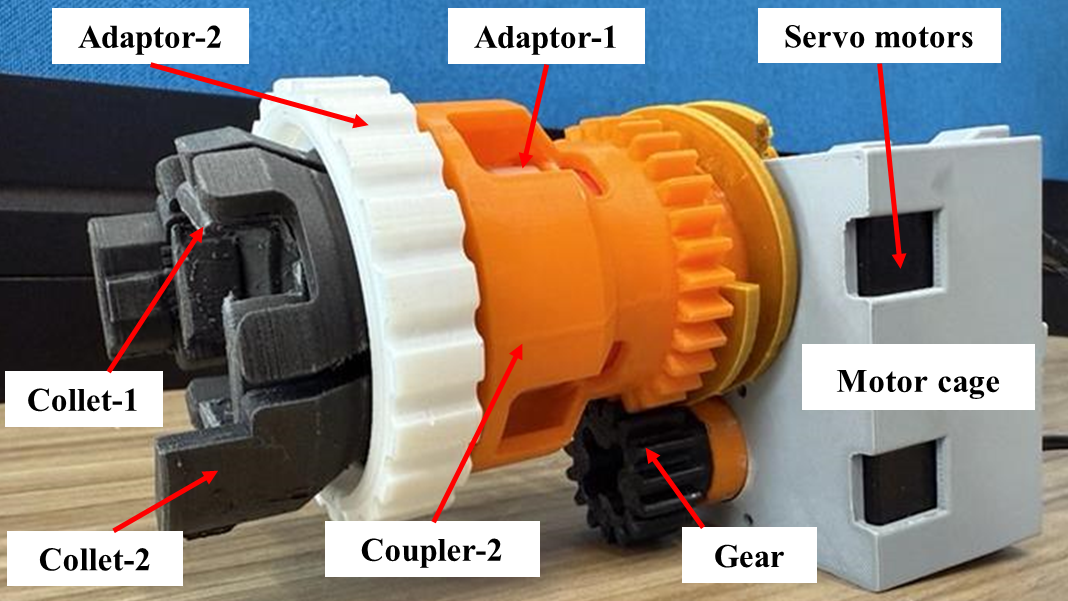}
    \vspace*{-7mm}
    \caption{Overview of the fabricated CH Gripping Mechanism, showcasing the 3D-printed components in their assembled configuration. Dimension are the same as exploded view.}
    \label{fig:Printed_bending_mechanism}
\end{figure}
We also analyze the design of the proposed mechanism by establishing a design space based on a mathematical modeling approach.
To thoroughly evaluate the functionality of the proposed design and mathematical framework in integration with a commercialized colonoscope (i.e., PENTAX EC-3840LK, PENTAX Medical), we performed  various simulation and experimental studies.    

\section{Design and Mathematical Modeling} \label{section:Design and Modeling}
\subsection{\textbf{Design Criteria}}

Our design objectives and criteria include proposing an effective and versatile mechatronics system that must satisfy the following features:  (i) \textit{Versatility} meaning that it should be readily usable with any existing commercialized endoscopic device regardless of their size differences and manufacturers; (ii) \textit{Simple  Mechanical Structure and Manufacturability} meaning that it should have a minimum number of components and motors required to actuate the three DoFs required during endoscopy while being easily manufacturable;
(iii) \textit{Seamless Integration} meaning that it should be easily integrated and disconnected with the CH of a colonoscope by clinicians; (iv) \textit{Intuitive User Interface} meaning that it should provide an intuitive user interface enabling simultaneous control of all DoFs without changing the existing surgical workflow.

\subsection{\textbf{Proposed Solution}}
Fig.\ref{fig:EasyC-Basic} illustrates the proposed modular mechatronics framework that satisfies all the aforementioned design criteria. 
As shown, this system  includes the following three main modules: (i) CH Gripping Mechanism, (ii) Feeder Mechanism, and (iii) User Interface. In the following sections, these modules are described in detail.
\indent\subsubsection{Design of CH Gripping Mechanism}
Inspired by the design of collet-chuck mechanisms commonly used in handheld drills and CNC lathes, we propose two concentric, modular, compact, and multi-jaw collet-chuck gripping mechanism to firmly hold the CH of the endoscope. This gripping mechanism simultaneously satisfies the above-mentioned design criteria and offers (i) a \textit{versatile design}, as its gripping size can readily be adjusted through the opening and closing of its jaws (shown in Fig.\ref{fig:BendingMechanism}a) by screwing/rotating adaptor over the collet's threaded surface. Thanks to the use of this mechanism, our proposed device can easily be integrated with various endoscopes/colonoscopes made by different companies that do not necessarily have identical geometries and dimensions; (ii) a \textit{simple design,} as the whole mechanism is made of only 5 main components making it very easy to fabricate and affordable (see Fig.\ref{fig:BendingMechanism}c); and (iii) a \textit{simple integration}, as the  user solely needs to sequentially assemble the concentric or nested collets and their adaptors (i.e., shown in Fig.\ref{fig:BendingMechanism}c). Of note, the components shown in Fig.\ref{fig:BendingMechanism}b are assembled once and remain in place, while to quickly connect/disconnect to the CH, only the components in Fig.\ref{fig:BendingMechanism}c need to be opened.

After firmly gripping the CH of endoscope by the concentric collets, these handles can independently and accurately be controlled using two motors (DYNAMIXEL XM540-W270-R, ROBOTIS, South Korea). Each motor is equipped with a 1:270 gear ratio and a 12-bit resolution absolute encoder.
Of note, as shown in Fig.\ref{fig:BendingMechanism}b, the outer collet is actuated through a 1:2 gear reduction mechanism, enabling precise bending in both left and right directions.

Figure~\ref{fig:BendingMechanism} shows in detail the geometry and dimensions of the designed components and the overall assembly of the gripping mechanism, based on the considered commercialized colonoscope (i.e., PENTAX EC-3840LK, PENTAX Medical). These dimensions are determined based on the mathematical framework described in the following section and are further refined to address additional considerations such as printability, potential part interference, and ergonomic factors related to the colonoscope’s CH. For instance, (i) the chuck at the collet’s end was enlarged to grip the control handle (CH) effectively and prevent saw-tip contact during small deflections, thereby enabling a wider gripping range; and (ii) four-leaf and two-leaf configurations were selected for the inner and outer collets, respectively, to accommodate the knob design of the selected colonoscope.

In this application, the material must provide both the flexibility needed for bending to securely grip the CH and the torsional strength required for efficient power transmission from the motor to the handle. After evaluating available 3D printing materials, we selected \textit{Markforged Onyx} (Markforged, Inc.) filament— a composite of nylon and chopped carbon fiber with a modulus of elasticity of $1.7 GPa$ (supplier value)— due to its high strength and stiffness. The part was printed horizontally using a \textit{Markforged Mark Two} 3D printer (Markforged, Inc.), as this orientation reduces the risk of collet failure by minimizing layer-to-layer weaknesses under combined torsional and bending loads. This choice ensures the component meets the functional requirements essential for optimal performance. Additionally, the remaining components were fabricated using Acrylonitrile Butadiene Styrene (ABS) on a Raise3D printer (Raise3D Technologies, Inc.) to provide the necessary rigidity for effective collet deformation, as shown in Fig.\ref{fig:Printed_bending_mechanism}.
\begin{figure*}[h!]
   \centering
    \includegraphics[width=2\columnwidth]{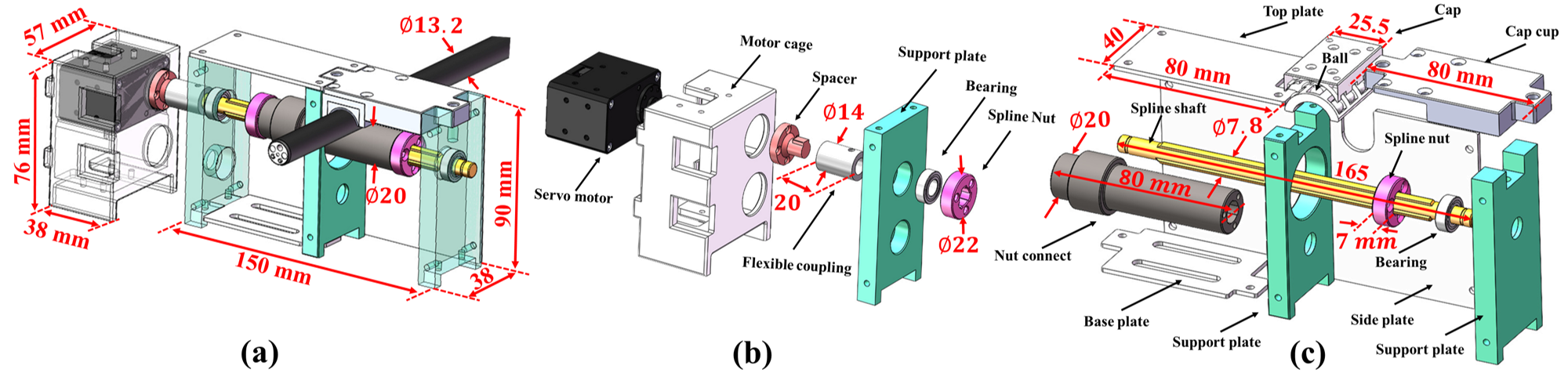}
    \vspace*{-4mm}
    \caption{Feeder Mechanism for insertion/retraction DoF: (a) Assembly view, and (b-c) Exploded view. All dimensions are in millimeter (mm).}
    \label{fig:InsertionMechanism}
\end{figure*}
\begin{figure}[t]
   \centering \includegraphics[width=1\columnwidth]{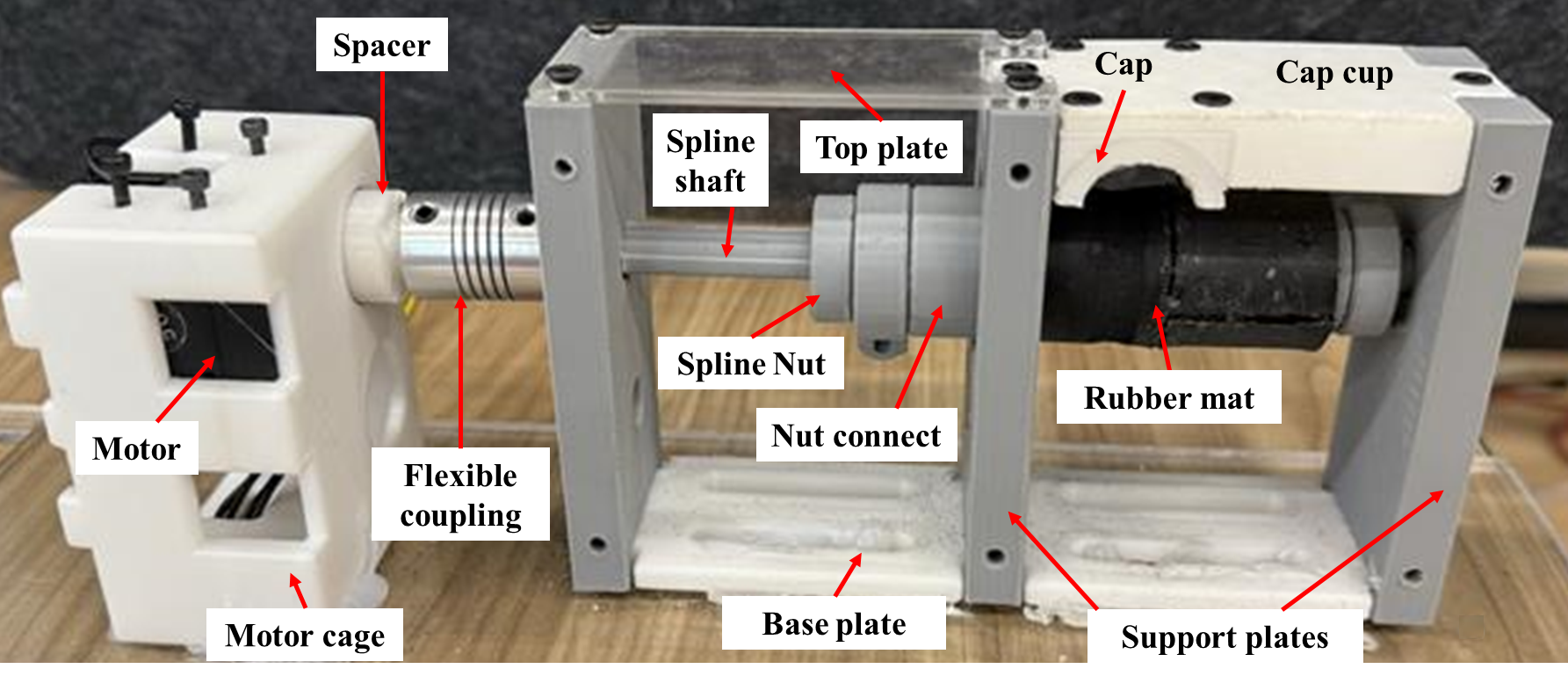}
    \vspace*{-7mm}
    \caption{Overview of the fabricated feeder mechanism, showcasing the 3D-printed components in their assembled configuration. Dimension are the same as exploded view.}
    \label{fig:Printed_feeder_mechanism}
\end{figure}
\indent\subsubsection{Design of Feeder Mechanism}
To meet the aforementioned criteria, the friction-based mechanism shown in Fig.\ref{fig:InsertionMechanism} was selected to control the insertion and retraction motions of a generic endoscope. This mechanism offers: (i) a \textit{versatile design}, allowing compatibility with various commercial colonoscopes of different sizes simply by loosening or tightening the cap screw; (ii) a \textit{simple design}, consisting of a minimal number of components; and (iii) \textit{seamless integration}, enabling clinicians to intuitively attach the mechanism to a generic endoscope without the need for assembly/disassembly —just by inserting/retracting the endoscope. 

After securely gripping the scope, insertion and retraction movements are controlled by a dedicated motor (DYNAMIXEL XM430-W350-R, ROBOTIS, South Korea). As shown in Fig.\ref{fig:InsertionMechanism}a, these movements are transmitted to the colonoscope through friction between the drive nut and the scope itself. The mechanism also incorporates a low-friction guide composed of idler balls (see Fig.\ref{fig:InsertionMechanism}c), which provide sufficient normal force —adjustable via preload tightening of screws—to enable the drive nut to advance the colonoscope, while still allowing it to slide freely. 
Although a thick layer of Urethane sleeves (rubber mat) was added to the spline shaft to improve the pushing force, the clinician can still adjust the maximum force by tightening or loosening the screws, thus preventing excessive pressure on the colon wall during insertion.
Furthermore, we designed the feeder mechanism, as shown in Fig.\ref{fig:Printed_feeder_mechanism}, independent of the bending mechanism to enable alignment with the patient without altering the configuration of the CH gripping mechanism.

\indent\subsubsection{User Interface}
A user interface has been implemented to allow users to intuitively control the mentioned Degrees of Freedom (DoFs) and steer the device. The current setup utilizes an XBOX $360\textsuperscript{TM}$ joystick (Microsoft Corporation) for mechanism control, as illustrated in Fig.\ref{fig:Experimental Setup}. This simple, intuitive, and affordable solution enables the user to simultaneously control and lock all three DoFs of an endoscopic device, taking pictures, and changing the illuminations by simply pushing a user-defined button. Of note, this user interface can effortlessly be swapped with other devices such as a haptic phantom or space mouse.
It is worth noting that alternative apparatus commonly used by surgeons to handle colonoscopes —such as jetting water to wash the colon surface and blowing air to inflate the colon for better maneuverability— can be seamlessly integrated. This design flexibility ensures that surgeons can operate the device with familiar and straightforward controls, promoting ease of use and efficiency in medical procedures.\\ 
To transmit commands from PC to DYNAMIXEL motor, we employ a U2D2 board (ROBOTIS Co., Ltd.) and a power hub. All motors are interconnected in a daisy chain configuration, facilitating communication with the setup via the PC. The Robot Operating System (ROS) is utilized as an open-source middleware framework for overall robot control. Furthermore, during maneuvering, data from the magnetic tracker is recorded, providing valuable information to determine the location and orientation of the end effector. This integrated approach allows for precise control and real-time data monitoring, enhancing the overall functionality and performance of the robotic system.

\subsection{\textbf{Mathematical Modeling of Collet}}
\begin{figure*}[h!]
    \centering
    \includegraphics[width=2\columnwidth]{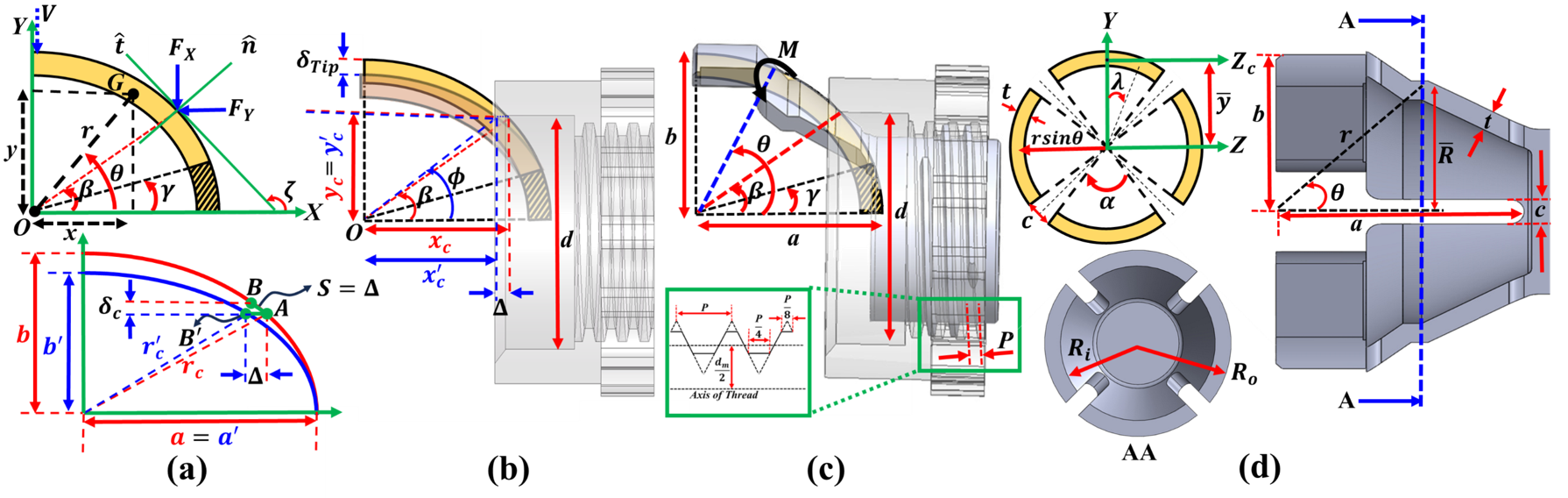}
    \vspace*{-5mm}
    \caption{ (a) Figure illustrates modeling of the collet with an elliptical arc to theoretically predict the tip deformation under the contact forces and virtual vertical force at the tip point; (b) Figure shows the beam configuration after applying a horizontal displacement at the contact point. Also, figures (c) and (D) demonstrate the primary parameters involved in modeling the collet.
    }
    \label{fig:collet model}
\end{figure*}

To establish a comprehensive mathematical relationship between the gripping behavior of the proposed collet (i.e., deflection of the collet jaws) with horizontal displacement of the adaptor (as the known input), in this section, we first propose a novel analytical modeling framework. Next, we  use this analysis to obtain a design space that assists us in designing a collet mechanism that can sufficiently grip the CH of an endoscope. 
As illustrated in Fig.\ref{fig:collet model}, we have modeled and simplified the symmetric jaws' geometry of the proposed collet as a cantilever beam  with a curved quarter ellipse. Figure compares the actual geometry of each jaw compared with the considered elliptic beam and shows the important parameters used for the modeling and design of the collet mechanism. As shown, these parameters include the semi-major axis (\(a\)) and semi-minor axis (\(b\)) of the elliptic arc, chuck width (\(c\)), inner diameter of adaptor (\(d\)), and thickness of collet (\(t\)).
Using these parameters and utilizing the Polar coordinate system shown in  Fig.\ref{fig:collet model}a, we can define the elliptic arc geometry of the jaws as follows: 
\begin{equation}
\begin{aligned}
\begin{cases}
    x = r(\theta) \cos\theta \\
    y = r(\theta) \sin\theta \\
    \frac{x^2}{a^2} + \frac{y^2}{b^2} = 1
\end{cases}
\Rightarrow r(\theta) = \frac{ab}{\sqrt{b^2 \cos^2\theta + a^2 \sin^2\theta}} \\
\end{aligned}
\label{eq:beta}
\end{equation}

where \(\theta\) and \(r(\theta)\) are the parameters describing the ellipse in the Polar coordinate system. Also, as shown in Fig.\ref{fig:collet model}a, \(x\) and \(y\) represent the location of a position of an arbitrary point (\(G\)) on the ellipse in the rectangular coordinate frame with the origin (\(O\)).  

The key step in our proposed modeling approach is to find the location of contact point between the collet and adaptor as the adaptor is screwed over the collet from its initial location $(x_c , y_c)$ with the horizontal displacement of $\Delta=0$. As illustrated in Fig.\ref{fig:collet model}a-b, through this course of motion, the initial contact location $A = (x_c , y_c)$ -- defined also by angle \(\beta\) and radius $r_c$ in the Polar coordinate -- will change to the next contact point $B' = (x_c', y_c')$ corresponding to the horizontal displacement \(\Delta\). As shown in Fig.\ref{fig:collet model}c, it is determined by the thread pitch \(P\) and number of revolutions of the adaptor around the collet \(N\) (\(\Delta = P N\)).
Given that the initial contact point $(x_c , y_c)$ between the adaptor and collet occurs at an angle \(\beta\), the vertical component of this initial contact point is equal to the inner radius of the adaptor (i.e., $y_c=d/2$, see Fig.\ref{fig:collet model}b). 
\begin{equation}
\begin{aligned}
     & y_{c} = r_c \sin \beta = \frac{ab}{\sqrt{b^2 \cos^2 \beta + a^2 \sin^2 \beta}} \sin\beta = \frac{d}{2} \\
     & \Rightarrow \frac{a^2 b^2 sin^2 \beta}{b^2 \cos^2 \beta + a^2 \sin^2 \beta} = \frac{d^2}{4} \Rightarrow \frac{a^2}{\cot^2 \beta + \frac{a^2}{b^2}} = \frac{d^2}{4}\\
     & \Rightarrow \cot\beta = \sqrt{\frac{4 a^2}{d^2}-\frac{a^2}{b^2}} \Rightarrow \beta = \cot^{-1}(\frac{a}{bd} \sqrt{4b^2-d^2})
\end{aligned}
\label{eq:contactpoint_initial}
\end{equation}
The next contact point $(x_c' , y_c')$  caused due to the horizontal movement of the adaptor (\(\Delta\)) can also be calculated as follows:

\begin{equation}
\begin{aligned}
\begin{cases}
    x'_c + \Delta = x_c \\
    x_c = r_c \cos\beta
\end{cases} \Rightarrow 
x'_c = \frac{ab \cos\beta}{\sqrt{b^2 \cos^2\beta + a^2 \sin^2\beta}} - \Delta
\end{aligned}
\label{eq:contactpoint_final}
\end{equation}

Since the cantilever beam's base angle and position remain unchanged, in our model, the major axis \(a\) of the ellipse stays constant, while the minor axis \(b\) is updated according to the deformation (as shown in Fig.\ref{fig:collet model}a). To determine the new minor axis \(b'\), we use the fact that the new contact point $(x'_c,\frac{d}{2})=(r_c \cos\beta - \Delta,\frac{d}{2})$ lies on the ellipse.

\begin{equation}
\begin{aligned}
    & \frac{x^2}{a^2}+\frac{y^2}{b'^2}=1 \xrightarrow{\begin{pmatrix} x'_c\\ \frac{d}{2} \end{pmatrix}} b' = \frac{ad}{2\sqrt{a^2-(r_{\beta} \cos \beta - \Delta)^2}}\\
\end{aligned}
\label{eq:updated b}
\end{equation}

Of note, here, we assume that the shape of the beam remains elliptical with the same semi-minor and major axes during the deformation and the collet does not experience a plastic deformation. To determine the location of the new contact point in the previous step, we use the fact that the adaptor slides on the surface of the collet, which has an elliptical shape. Consequently, the arc length between the two contact points (i.e., points \(A\) and \(B\)) on the surface of the ellipse in the previous step is equal to \(\Delta\), as shown in Fig.\ref{fig:collet model}a. Based on this knowledge, the angle of point \(B\) can be determined by satisfying the following equation:
\begin{equation}
\begin{aligned}
    \int_{\beta}^{\phi} \sqrt{r^2 + (\frac{dr}{d\theta})^2} \, d\theta = \int_{\beta}^{\phi} \frac{ab}{\sqrt{b^2 \cos^2\theta + a^2 \sin^2\theta}} \, d\theta = \Delta
\end{aligned}
\label{eq:Next angle}
\end{equation}
where \(\phi\) represents the angle corresponding to the new contact point on the previous ellipse (i.e., point \(B\)) in polar coordinate. Of note, calculating \(\phi\) requires a numerical and iterative approach (i.e., Bisection or Newton–Raphson methods \cite{Chapra2001NumericalMF}).

After finding the relationship between the adaptor's movement \(\Delta\) and the contact point between the adaptor and collet, we are able to calculate the deflection behavior \(\delta_{Tip}\) of the collet's jaws as the adaptor moves. Since the interaction force between the collet and adaptor is unknown and variable as the adaptor is screwed over the collet, the Castigliano's method  \cite{HEARN1997254} has been utilized.  Castigliano's second theorem, as a powerful energy-based technique, relies on the principle that the first partial derivative of the total strain energy (internal energy) with respect to the force applied at any point is equal to the deflection at this point in the direction of the force line. \\
If one is interested in obtaining the deflection of a point in a specific direction where no actual force is applied on that point, a virtual force first needs to be applied at this point parallel to the deformation direction of interest. After the analysis, this virtual force is substituted with zero in the obtained equations. Following this procedure, to determine the vertical tip deflection of the collet jaws as well as the deflection of contact points. As shown in Fig.\ref{fig:collet model}a, we first introduced the virtual vertical force $V$ at the collet tip, as well as vertical and horizontal forces $F_X$ and $F_Y$ at the contact point of collet and adaptor. Although the interaction force between the collet and adaptor is unknown at the contact point, using (\ref{eq:contactpoint_initial}), we have geometric knowledge about the vertical deformation of the collet at this point (\( r(\phi) \sin \phi - \frac{d}{2} \)). Therefore, using Castigliano's method, we can determine the deformation of the contact point by applying unknown vertical and horizontal forces. Of note, $F_X$ and $F_Y$ are treated as independent variables in Castigliano's approach. This choice is made to achieve the vertical displacement independently. By solving the resulting equations, we can establish a relationship between the unknown forces acting at the contact point.
Therefore, we have derived the momentum at each cross-sectional area of the ellipse. Given the presence of a virtual force at the tip and both vertical and horizontal forces at the contact point, we can distinguish between two distinct regions: from the base to the contact point and from the contact point to the tip. In the framework, where $F_X$ and $F_Y$ are represented, we aim to calculate the momentum of each force. To achieve this, we determine the momentum arm of each force using polar representation. It is important to note that our focus is on the end part of the beam, specifically from the cross-sectional point \(\theta\) to the tip point, as illustrated in Fig.\ref{fig:collet model}c. Because there is no need to calculate the reaction force at the base in this context.
\begin{equation}
\begin{aligned}
   M^{\star} &= F_Y (r_{\theta} \cos\theta - r_{\beta} \cos\beta) + F_X (r_{\beta} \sin\beta - r_{\theta} \sin\theta) \\
    M &= 
    \begin{cases}
       V r(\theta) \cos\theta + M^{\star}, & \text{if } \gamma \leq \theta \leq \beta \\
       V r(\theta) \cos\theta, & \text{if } \beta < \theta \leq \frac{\pi}{2}
    \end{cases} \\
\end{aligned}
\label{eq:Momentum}
\end{equation}

Subsequently, we proceed to integrate along the ellipse curve to obtain the beam's energy. Due to the independence V, $F_X$, and $F_Y$ from \(\theta\), we employ Leibniz's integral rule to simplify Castigliano's method for obtaining the displacement equations.
 
\begin{equation}
\begin{aligned}
    &\delta_{c} = \int_{\gamma}^{\frac{\pi}{2}} \frac{M}{EI} \frac{\partial M}{\partial F_y} \,r(\theta) d\theta =  r(\phi) \sin\phi - \frac{d}{2} \\
    &\delta_{Tip} = \int_{\gamma}^{\frac{\pi}{2}}\frac{M}{EI} \frac{\partial M}{\partial V} \,r(\theta) d\theta \\
\end{aligned}
\label{eq:Castigliano}
\end{equation}
where \(M\), \(E\), and \(I\) represent the internal moment at a point, Young's modulus, and the second moment of area of the cross-section around neutral axis, respectively. Also, \(\delta_{c}\)\ and \(\delta_{Tip}\)\ denote vertical deflection of contact point and tip, respectively. Since the cross-section of a saw is a circular arc with a variable radius along the elliptical shape, as shown in Fig.\ref{fig:collet model}d, the second moment of area in \ref{eq:Castigliano} varies. To calculate the second moment of area around the neutral axis (center of the area), we first determine it around the \(Z\) axis. 
\begin{equation}
\begin{aligned}
     & I_{ZZ} = \iint y^2 dx dy = \int_{R_i}^{R_o}\int_{-\alpha/2}^{\alpha/2} (R cos\lambda)^2 R d\lambda dR = \\
     & \int_{R_i}^{R_o} R^3 dR \int_{-\alpha/2}^{\alpha/2} \frac{1+cos 2 \lambda}{2} d\lambda = \frac{R^4_o-R^4_i}{4} \frac{\alpha + sin \alpha}{2} \Rightarrow \\ 
     & I_{ZZ} = \underbrace{(R^4_o-R^4_i)}_{(R^2_o+R^2_i)(R_o+R_i)(R_o-R_i)} \frac{\alpha + sin \alpha}{8} \approx \overline{R}^3t\frac{\alpha + sin \alpha}{2}
\end{aligned}
\label{eq:Second moment of area}
\end{equation}
where \(R_o\), \(R_i\), and \(\overline{R}\) represent outer, inner, and mean  radius of circular arc, respectively. Also, \(\alpha\)\ and \(t\)\ denote central angle and saw thickness. Of note, the central angle varies along the ellipse and depends on both the chuck size and the mean radius, which itself varies along the ellipse, as follows:
\begin{equation}
\begin{aligned}
     &\alpha = \frac{\pi}{2} - 2 sin^{-1}(\frac{c}{2 \overline{R}}) 
\end{aligned}
\label{eq:Second moment of area}
\end{equation}
Then, using the parallel axis theorem—which states that the second moment of area around any parallel axis is equal to the second moment of area around the centroidal axis plus the product of the area and the square of the distance between the two axes—we transfer it to the centroidal axis \(Z_c\).
\begin{equation}
\begin{aligned}
    \begin{cases}
        I_{ZZ} = I_{Z_c Z_c} + A \overline{y}^2 \\
        \overline{y} = \frac{\int y dx dy}{A} = \frac{\int_{-\alpha/2}^{\alpha/2} \overline{R} cos\lambda \overline{R} t d\lambda}{\overline{R} \alpha t} = \frac{2 \overline{R} sin \frac{\alpha}{2}}{\alpha}
     \end{cases} \\
     \Rightarrow I = I_{Z_c Z_c} = \overline{R}^3t(\frac{\alpha + sin \alpha}{2} - 4 \frac{sin^2 \frac{\alpha}{2}}{\alpha})
\end{aligned}
\label{eq:Parallel axis theorem}
\end{equation}

where \(\overline{R}\) is equal to \(r sin \theta\) in this context, as shown in Fig.\ref{fig:collet model}d.

\indent Also, the first derivatives of the internal moment with respect to \(F_y\) and \(V\), $\frac{\partial M}{\partial F_y}$ and $\frac{\partial M}{\partial V}$, are as follows:

\begin{equation}
\begin{aligned}
    \frac{\partial M}{\partial F_y} &= 
    \begin{cases}
       r(\theta) \cos\theta - r(\beta) \cos\beta, & \text{if } \gamma \leq \theta \leq \beta \\
       0, & \text{if } \beta < \theta \leq \frac{\pi}{2}
    \end{cases} \\
    \frac{\partial M}{\partial V} &= 
    \begin{cases}
       r(\theta) \cos\theta, & \text{if } \gamma \leq \theta \leq \beta \\
       r(\theta) \cos\theta, & \text{if } \beta < \theta \leq \frac{\pi}{2}
    \end{cases} \\
\end{aligned}
\label{eq:First derivation of Momentum}
\end{equation}

As mentioned previously, in partial differentiation, horizontal (\(F_X\)) and vertical (\(F_Y\)) forces at the contact point are assumed to be independent. However, they are inherently related, since the total force must be normal to the surface. This relationship can be established by ensuring that the tangential components of these forces are zero. To achieve this, the normal ($\hat{n}$) and tangential ($\hat{t}$) coordinates are defined at the contact point, as illustrated in Fig.\ref{fig:collet model}a. The relationship between these forces is given by:
\begin{equation}
\begin{aligned}
        \frac{F_Y}{F_X} = \tan\left(\zeta-\frac{\pi}{2}\right) = -\cot\zeta \\
\end{aligned}
\label{eq:Contact Force}
\end{equation}

where \(\zeta\) represents the angle between the tangent line at the contact point of the ellipse and the horizontal axis.
Since tangent of this angle (\(\tan \zeta\)) is equivalent to the derivative of y with respect to x (\(\frac{dy}{dx}\)) at contact point, the first derivative of the ellipse equation in Cartesian coordinates is obtained:
\begin{equation}
\begin{aligned}
\begin{cases}
    \tan\zeta = \frac{dy}{dx} \\
    \frac{2x_{c}}{a^2} + \frac{2y_{c}}{b^2}  \frac{dy}{dx} = 0
\end{cases}
\Rightarrow \tan\zeta = -\frac{b^2}{a^2} \frac{x_c}{y_c} 
\end{aligned}
\label{eq:Geometry}
\end{equation}

Given the connection between $y_c$ and $x_c$ in the Cartesian coordinate with $\beta$ in the polar coordinate, Eq.~\ref{eq:Geometry} can be simplified.  
\begin{equation}
\begin{aligned}
\begin{cases}
    \tan\zeta = -\frac{b^2}{a^2} \frac{x_c}{y_c}\\
    \frac{x_c}{y_c} = \frac{r_{\beta} cos\beta}{r_{\beta} sin\beta} = \cot\beta
\end{cases}
\Rightarrow \tan\zeta = -\frac{b^2}{a^2} \cot\beta 
\end{aligned}
\label{eq:Geometry_simplified}
\end{equation}
Using Eq.~\ref{eq:Contact Force} and ~\ref{eq:Geometry_simplified}, the relationship between the contact forces and the angle of the contact force can be determined.

\begin{equation}
\begin{aligned}
    \frac{F_Y}{F_X} = \frac{a^2}{b^2} \tan\beta
\end{aligned}
\label{eq:Contact Force concluded}
\end{equation}

Since the vertical force \(V\) at the collet tip is virtual, it substitutes by zero after first derivation. Consequently, the momentum, as stated in \ref{eq:Momentum}, will be zero in the interval \((\beta,\frac{\pi}{2}]\), and integral in \ref{eq:Castigliano} will be simplified over the interval \([\gamma,\beta]\).

\begin{equation}
\begin{aligned}
    & \int_{\gamma}^{\beta} \frac{M\Big|_{v=0}}{EI} \frac{\partial M}{\partial F_y} \,r_{\theta} d\theta =  r(\phi) \sin\phi - \frac{d}{2} \Rightarrow  \\ 
    & F_X \underbrace{\int_{\gamma}^{\beta} \frac{(r_{\beta} \sin\beta - r_{\theta} \sin\theta)(r_{\theta} \cos\theta - r_{\beta} \cos\beta) r_{\theta}}{E (r_{\theta} sin \theta)^3 t(\frac{\alpha + sin \alpha}{2} - \frac{4sin^2 \frac{\alpha}{2}}{\alpha})} d\theta}_{a_x} + \\
    & F_Y \underbrace{\int_{\gamma}^{\beta} \frac{(r_{\theta} \cos\theta - r_{\beta} \cos\beta)^2 r_{\theta}}{E (r_{\theta} sin \theta)^3 t(\frac{\alpha + sin \alpha}{2} - 4 \frac{sin^2 \frac{\alpha}{2}}{\alpha})} d\theta}_{a_y} = r(\phi) \sin\phi - \frac{d}{2} \\ 
\end{aligned}
\label{eq:Simplified Castigliano}
\end{equation}

These parameters defined within braces (i.e., \(a_x\) and \(a_y\)) are constant values for each fixed geometry. 
By solving the system of linear equations (i.e., Eq.\ref{eq:Contact Force concluded} and Eq.\ref{eq:Simplified Castigliano}) in two variables (i.e., \(F_X\) and \(F_Y\)), these unknown parameters are calculated.   
\begin{equation}
\begin{aligned}
    & F_X = \frac{r(\phi) \sin\phi - \frac{d}{2}}{a_x+a_y \frac{a^2}{b^2} \tan\beta} \\
    & F_Y = \frac{a^2}{b^2} \tan\beta \frac{r(\phi) \sin\phi - \frac{d}{2}}{a_x+a_y \frac{a^2}{b^2} \tan\beta} = \frac{r(\phi) \sin\phi - \frac{d}{2}}{a_x \frac{b^2}{a^2} \cot\beta+a_y}
\end{aligned}
\label{eq:Solve}
\end{equation}

With these concluded values, the tip deflection can be computed using the second equation of Eq.\ref{eq:Castigliano}. 
\begin{equation}
\begin{aligned}
    & \delta_{Tip} = \int_{\gamma}^{\frac{\pi}{2}}\frac{M\Big|_{v=0}}{EI} \frac{\partial M}{\partial V} \,r(\theta) d\theta \Rightarrow\\
    & \delta_{Tip} = F_X \underbrace{\int_{\gamma}^{\beta} \frac{(r_{\beta} sin\beta - r_{\theta} sin\theta) r^2_{\theta} cos\theta}{E (r_{\theta} sin \theta)^3 t(\frac{\alpha + sin \alpha}{2} - 4 \frac{sin^2 \frac{\alpha}{2}}{\alpha})} d\theta}_{b_x} \\
    & + F_Y \underbrace{\int_{\gamma}^{\beta} \frac{(r_{\theta} cos\theta - r_{\beta} cos\beta) r^2_{\theta} cos\theta}{E (r_{\theta} sin \theta)^3 t(\frac{\alpha + sin \alpha}{2} - 4 \frac{sin^2 \frac{\alpha}{2}}{\alpha})} d\theta}_{b_y} \\ 
\end{aligned}
\label{eq:Tip displacement}
\end{equation}
By substituting the \ref{eq:Solve} into \ref{eq:Tip displacement},  the relationship between the vertical tip displacement and the vertical contact displacement is derived.  
\begin{equation}
\begin{aligned}
\delta_{Tip} = \frac{b_x + b_y \frac{a^2}{b^2} \tan\beta}{a_x+a_y \frac{a^2}{b^2} \tan\beta} (r(\phi) \sin\phi - \frac{d}{2})
\end{aligned}
\label{eq:relation Tip-Contact}
\end{equation}
\begin{algorithm}[t]
\label{alg:algorithm1}
	\caption{Solving Tip Displacement Analytically}
\SetAlgoLined

\nonl\hl{\textbf{Input:}}  $a$, $b$, $c$, $d$, $\gamma$, $\Delta$, $t$, $E$;

\nonl\textbf{Initialization:}

$n$ $\longleftarrow$ Iteration number to reach out final horizontal movement (i.e., 60);\\

$k \longleftarrow 0$; \\

$\delta^{0} \longleftarrow 0$;

\nonl\textbf{Iterative Process:}

\Do{$k \neq n$ }{ $\beta$ $\longleftarrow$ Calculate contact angle using (\ref{eq:contactpoint_initial}) [\(a\), \(b\), \(d\)];\\

$\phi \longleftarrow$ 
Solve (\ref{eq:Next angle}) using the bisection method; \\
$\delta^{k+1} \longleftarrow \delta^{k} + $ Calculating the tip displacement using (\ref{eq:relation Tip-Contact}) [$a$, $b$, $c$, $d$, $\phi$, $\beta$];

$b \longleftarrow$ Updating minor-axis of ellipse after deformation using (\ref{eq:updated b}) [$a$, $b$, $d$, $\beta$, $\frac{\Delta}{n}$]; \\

$k \longleftarrow k + 1$; \\
}

\nonl\textbf{Output:} $\delta$

\end{algorithm}

Since saw thickness \(t\) and elastic modulus \(E\) remain constant along the ellipse, these parameters pass through the integral and cancel out in \ref{eq:relation Tip-Contact}. Consequently, although the contact force correlates with the saw thickness and elastic modulus, the tip displacement —given the vertical displacement of contact point— becomes independent of these parameters. 

\section{Evaluation Studies and Results}
\subsection {\textbf{Simulation Studies Using Analytical Model}}

Upon completion of the model in \textit{SOLIDWORKS} (Dassault Systèmes), a \textit{static analysis} is carried out as part of the design process. This analysis determines the deformation of the collets under the influence of the adaptor's tightening (\(\Delta\)), aiming to identify the minimum knob diameter that the collet can securely grip. However, due to the high computational cost, time consumption, and complexity, this approach is impractical. Instead, using the proposed mathematical model provides a practical alternative for generating the design space, conducting sensitivity analysis for each parameter, and selecting the optimal parameters based on our requirements.

\begin{figure}[t]
   \centering
   \includegraphics[width=1\columnwidth]{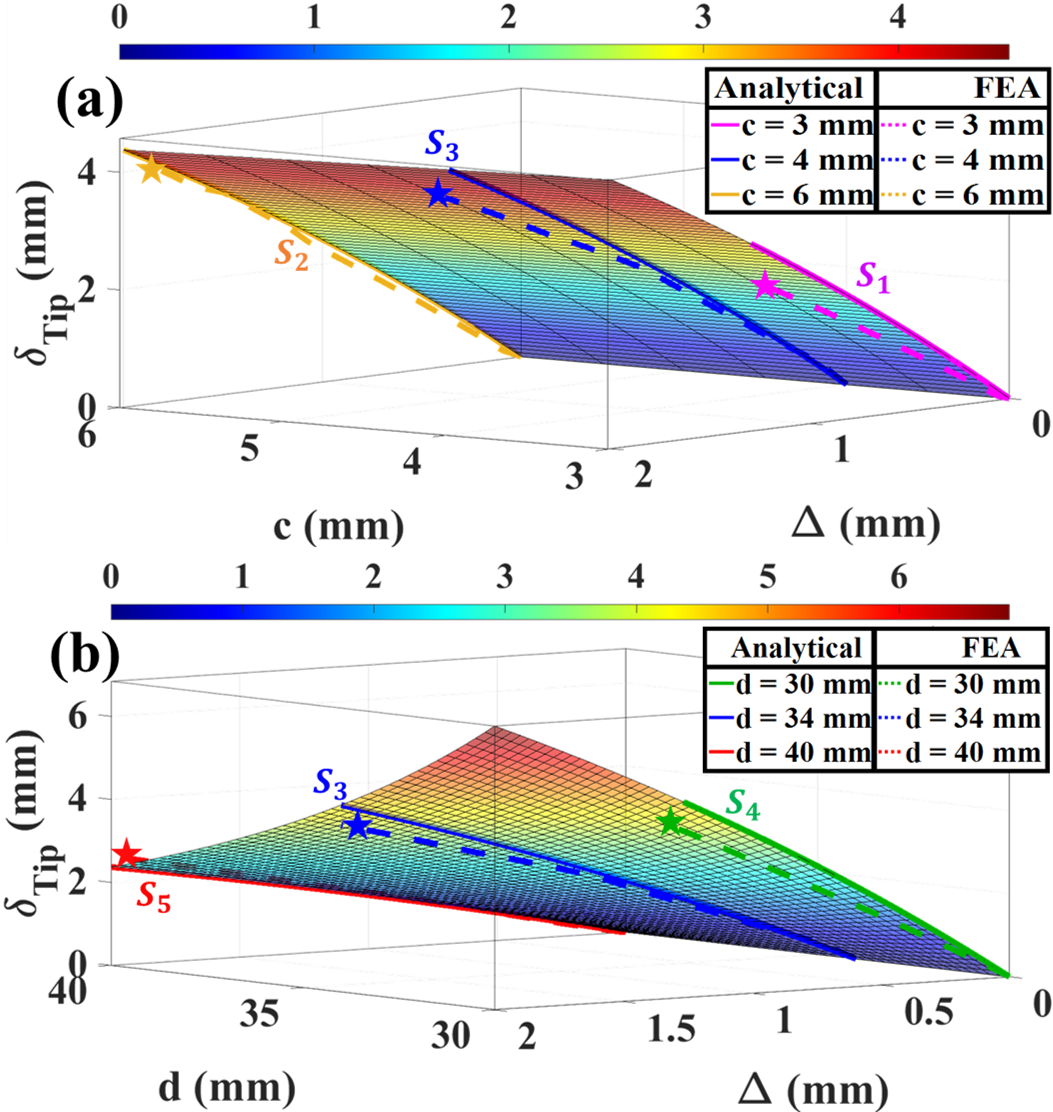}
    \vspace*{-7mm}
    \caption{The design space, along with the S1–S5 design curves, shows the relationship between collet deformation and adaptor horizontal movement for (a) various chuck sizes (scenario I) and (b) various adaptor diameters (scenario II). The solid lines depict the analytical collet deformation, whereas the dashed lines represent the collet deformation obtained from FEA.}
    \label{fig:workspace}
\end{figure}

By employing a recursive approach, as summarized in Algorithm \ref{alg:algorithm1}, the design space of the collet is generated. Choosing the smallest increment size (i.e., $\frac{\Delta}{n}$) yields more accurate results for the tip displacement in each horizontal movement of the adaptor. Considering a 1.5 mm clearance between the knob and collet on each side to meet quick assembly requirement, the horizontal movement continues until the tip displacement reaches this value. Of note, once the collet reaches the knob in this deformation, further tightening of the adaptor by rotation does not result in additional tip displacement but increases the gripping force. With this assumption, the “design space” obtained by written code in \textit{MATLAB} (MathWorks, Inc.) is illustrated in Fig.\ref{fig:workspace} by varying the geometrical parameters (i.e., chuck size and adaptor diameter representing scenario I and scenario II, respectively) to clearly demonstrate their effects on tip displacement. This sensitivity analysis provides a variety of geometries to achieve the desired collet closer size, ensuring secure gripping of the knob while meeting additional requirements such as avoiding interference between nested collets.

\begin{table}[t]
    \centering
    \caption{Parameters of the chosen samples for evaluating the analytical outcome by FEA, along with the maximum and mean absolute errors between the analytical and FEA results.}
    \vspace*{-2mm}
    \label{Samples}
    \resizebox{\columnwidth}{!}{%
        \begin{tabular}{ccccccc}
            \toprule
            \toprule
            \textbf{Sample} & \textbf{a} & \textbf{b} & \textbf{c} & \textbf{d} & \textbf{ME} & \textbf{MAE} \\
            & \textbf{(mm)} & \textbf{(mm)} & \textbf{(mm)} & \textbf{(mm)} & \textbf{(mm)} & \textbf{(mm)} \\
            \midrule
            $S_1$ & 34 & 26.5 & 3  & 34 & 0.60 & 0.36 \\
            $S_2$ & 34 & 26.5 & 6  & 34 & 0.17 & 0.11\\
            $S_3$ & 34 & 26.5 & 4  & 34 & 0.49 & 0.21\\
            $S_4$ & 34 & 26.5 & 4  & 30 & 0.72 & 0.44\\
            $S_5$ & 34 & 26.5 & 4  & 40 & 0.26 & 0.09\\
            \bottomrule
        \end{tabular}%
    }
\end{table}
\subsection{\textbf{Simulation Studies Using Finite Element Analysis (FEA)}}
 To evaluate the results obtained from the analytical model, we performed static analysis using \textit{ABAQUS}  (Dassault Systèmes) software for five samples (i.e., $S_1  \thicksim S_5$) shown in Fig.\ref{fig:workspace} and summarized in Table \ref{Samples}. Of note, due to the computational cost of validating the whole design space, we compared the outcomes of several extreme cases from the analytical model with those from a Finite Element Analysis (FEA). In the performed studies, we incorporated the following strategies to enhance the accuracy and stability of the FEA analysis: (i) a frictionless contact model between the adaptor and collet to isolate contact interactions, (ii) a solid adaptor to eliminate the effects of bearing stress, ensuring a more controlled evaluation, and (iii) smooth amplitude curves (i.e., splines) for horizontal movement to mitigate numerical instabilities and prevent abrupt load variations that could lead to convergence issues. For meshing, we utilized a linear tetrahedral mesh with a global element size of $1$ mm and $0.75$ mm for the collet and adaptor, respectively. Markforged Onyx (Markforged, Inc.)  and ABS also were selected as the materials, with Young’s moduli of $1.7$ GPa (supplier value, Markforged, Inc.) and $2.2$ GPa (supplier value, Raise3D Technologies, Inc.) for the collet and adaptor, respectively. We employed the \textit{direct} equation solver with the \textit{full Newton} solution technique to perform the FEA in  \textit{ABAQUS}.

\begin{figure}[t]
   \centering
   \includegraphics[width=1\columnwidth]{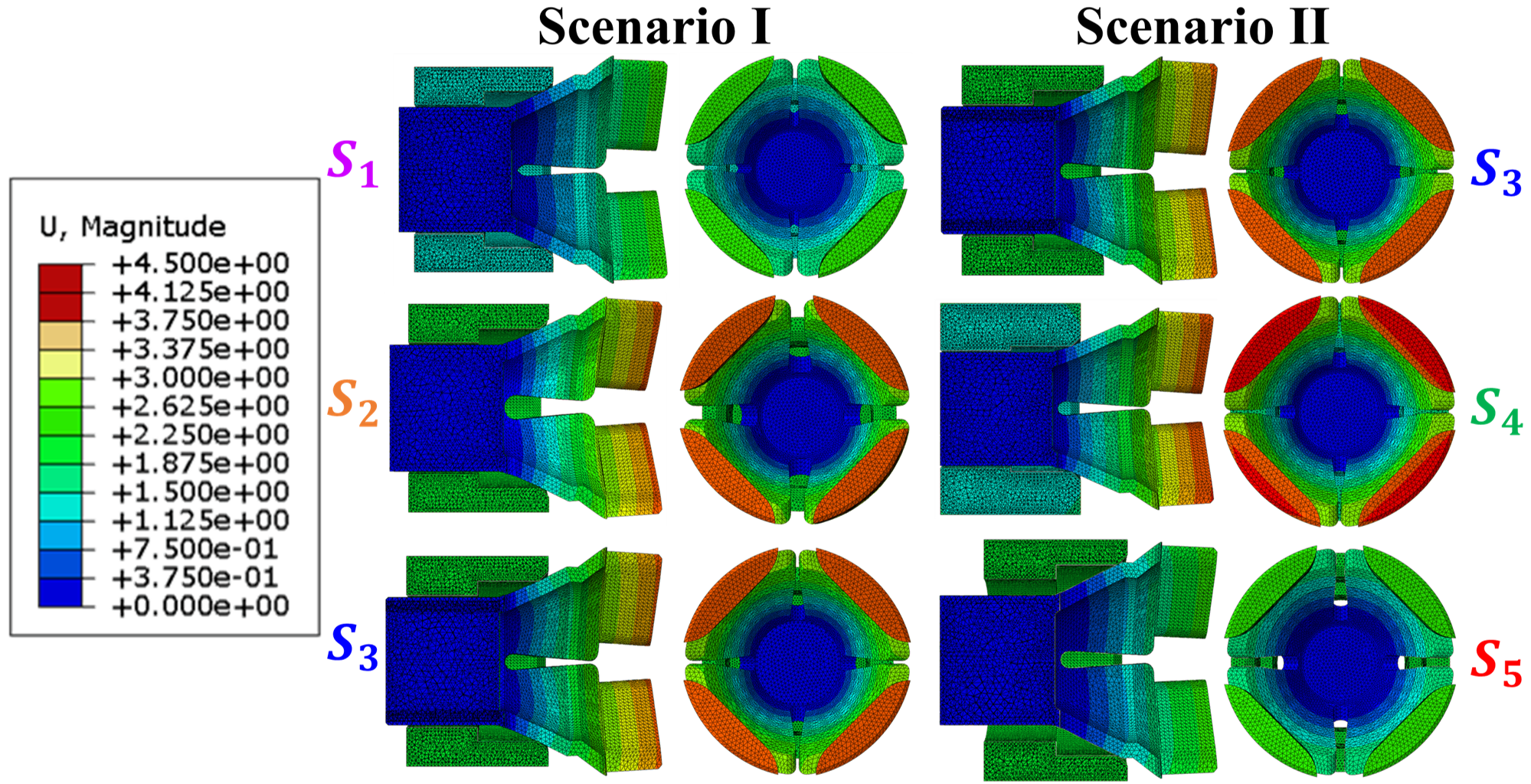}
    \vspace*{-7mm}
    \caption{ Simulation results of collet deformation for five different collet and adaptor geometries under an applied input in two different scenarios.}
    \label{fig:FEA_Scenario}
\end{figure}
Figure \ref{fig:workspace} shows the “\textit{design space}” together with a comparison of the “\textit{design curves}” associated with $S1 \thicksim S5$ results from the analytical model and those from the \textit{ABAQUS} simulations. Here, “\textit{design space}” is generated by Algorithm \ref{alg:algorithm1}, demonstrating the deformation of collet tip with all possible chuck size (Scenario I) and adaptor diameter (Scenario II) for a range of horizontal adaptor movements. 
In this case, the aforementioned “\textit{design space}” will be reduced to a “\textit{design curves}” for extreme cases of chuck size and adaptor diameter to compare accuracy and reliability between the analytical model and FEA simulation in capturing the deformation behavior of the system in operational region. The maximum error (ME) and mean absolute error (MAE) between result of analytical model and FEA for these five cases are summarized in Table.\ref{Samples}. Figure. \ref{fig:FEA_Scenario} also shows both the front view and the cross-sectional side view of the collet in the final stage for all cases (marked by a "$\star$" sign in Fig.\ref{fig:workspace}), which is obtained by FEA using \textit{ABAQUS}.

\subsection{\textbf{Experimental Studies}}
\subsubsection{Experimental Setup} 
\begin{figure}[t]
   \centering
   \includegraphics[width=1\columnwidth]{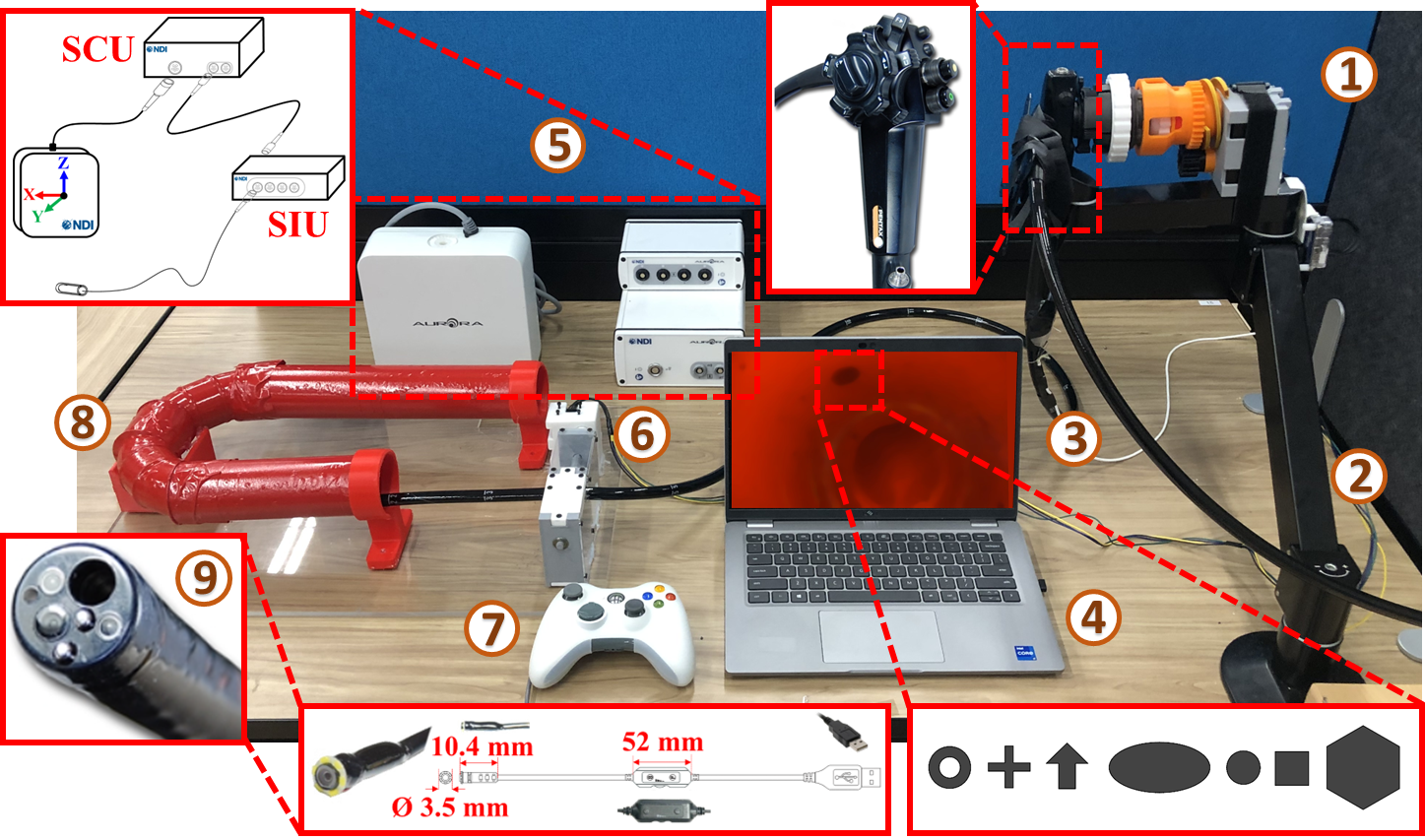}
    \vspace*{-8mm}
    \caption{Experimental Setup used for performing the case study, including: \raisebox{.5pt}{\textcircled{\raisebox{-.9pt} {1}}} - CH Gripping Mechanism placed on a rotary stand, \raisebox{.5pt}{\textcircled{\raisebox{-.9pt} {2}}} - A passive rotary stand, \raisebox{.5pt}{\textcircled{\raisebox{-.9pt} {3}}} - PENTAX EC-3840LK colonoscope, \raisebox{.5pt}{\textcircled{\raisebox{-.9pt} {4}}} - Monitoring the interior of tube to detect markers, \raisebox{.5pt}{\textcircled{\raisebox{-.9pt} {5}}} - Aurora NDI magnetic tracker, \raisebox{.5pt}{\textcircled{\raisebox{-.9pt} {6}}} - Feeder Mechanism, \raisebox{.5pt}{\textcircled{\raisebox{-.9pt} {7}}} - XBOX $360\textsuperscript{TM}$ joystick controller, \raisebox{.5pt}{\textcircled{\raisebox{-.9pt} {8}}} - The U-shaped tube with various markers are attached on its internal surface, and \raisebox{.5pt}{\textcircled{\raisebox{-.9pt} {9}}} - The end effector of colonoscope includes a MISUMI XD-VB164A03LH-120 camera.}
    \label{fig:Experimental Setup}
\end{figure}

Figure \ref{fig:Experimental Setup} shows the experimental  setup used for performing the experiments, including the following components: (1) The  CH gripping mechanism designed to control the bending motion of the colonoscope (left-right and top-down) and steer it within the phantom; (2) A passive rotary stand that facilitates easy alignment of the setup with the patient; (3) A commercial colonoscope (PENTAX EC-3840LK colonoscope); (4) A PC used for system control and real-time visualization of the phantom (represented here by a U-shaped tube); (5) An Aurora NDI magnetic tracker, comprising the Planar Field Generator 20-20 FG, the System Control Unit (SCU), and the Sensor Interface Unit (SIU) for precise spatial tracking of colonoscope tip. The magnetic tracker operates by measuring induced current in the sensor via the SIU in a uniformly generated field by the Planar Field Generator. The SCU calculates position and orientation based on these parameters at $40 Hz$ sampling rate, transferring digitized data to the computer through a USB connection; (6) The proposed Feeder mechanism module to enables insertion and retraction of the colonoscope, (7) XBOX 360 joystick controller serves as a user interface for operating the robotic colonoscope using assigned button functions (i.e., Button \textbf{$B_L$} – controls insertion of the scope to reach markers using feeder mechanism, Button \textbf{$B_R$} – controls retraction of the scope for removal using feeder mechanism, and Axis \textbf{$J_L$} – Adjusts the orientation of the colonoscope tip using CH gripping mechanism; (8) A U-shaped tube phantom which various markers are attached inside it for identification during internal examinations,(9) A MISUMI XD-VB164A03LH-120 camera (MISUMI Electronics Corp.) to capture images from phantom, measuring Ø3.5 x 10.4 mm, featuring a USB Type-A interface, a dimmer for adjusting illumination, and S/F buttons for image freezing.
The adoption of this camera eliminates the need for an expensive power switch on the colonoscope, streamlining image and illumination provision.

\subsubsection{{Software Architecture}}
\begin{figure}[t]
   \centering
    \includegraphics[width=1\columnwidth]{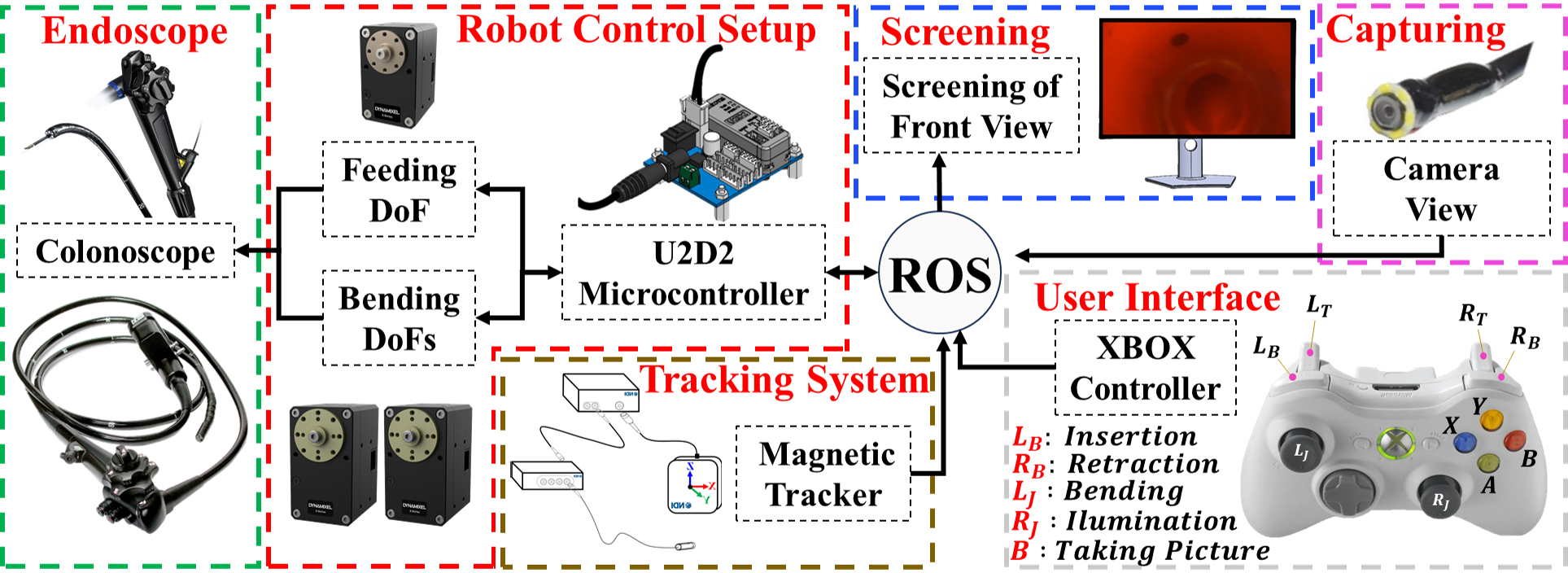}
    \vspace*{-7mm}
    \caption{The block diagram of the control system and software architecture in the ROS environment consists of distinct modules.}
    \label{fig:BlockDiagram}
\end{figure}

The control system and data collection during colonoscopy procedures are implemented within the Robot Operating System (ROS) environment to ensure real-time operation and synchronized data acquisition. As illustrated in Fig.\ref{fig:BlockDiagram}, ROS acts as a supervisory framework, enabling seamless and lag-free communication between different functional blocks or nodes. For instance, the \textit{robot control setup} (Red box) represents the proposed robotic system that interfaces with the \textit{endoscopic tool} (Green box) through bidirectional communication—subscribing to user commands and publishing encoder data to the ROS environment. The \textit{tracking system} (Brown box) provides unidirectional communication by publishing the position and orientation of the endoscope tip. The \textit{user interface} block (Gray box) is responsible for publishing the user's commands to ROS. The \textit{capturing} block (Pink box) delivers live video streaming, which is displayed to the surgeon via the \textit{screening} block (Blue box). Also, whole data of procedure (i.e., position and orientation of colonoscope tip by NDI magnetic tracker, input command by joystick, and encoders values) is recorded using ROS bag packages to perform automatic maneuvering or guiding for follow-up inspections, precisely revisiting the location where polyps were removed after a six-month interval. This comprehensive setup, coupled with the sensors, creates an environment where surgeons can solely focus on identifying polyps and cancer without distraction.
Moreover, this setup offers valuable support for training senior doctors. The maneuvering and detection skills of users can be assessed by analyzing encoder and magnetic tracker data, allowing for comparisons not only with other individuals but also with the user's own performance over time. This quantitative analysis, based on specifications or mathematical parameters, enhances the training evaluation process.
\subsubsection{{Experimental Procedure}}
\begin{figure}[t]
   \centering
    \includegraphics[width=1\columnwidth]{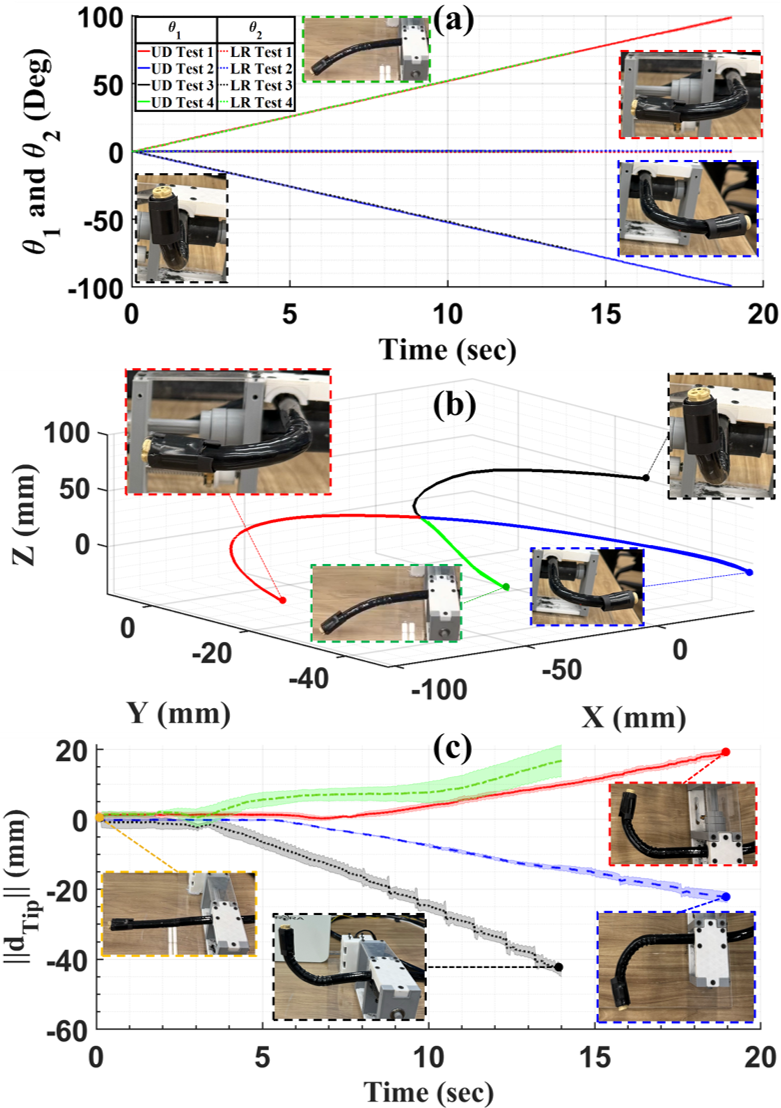}
    \vspace*{-7mm}
    \caption{ Movements of the colonoscope tip measured by a magnetic tracker in various tests, including bending to the right (Red), left (Blue), down (Green), and up (Black) using CH gripping mechanism. The graphs depict: (a) Encoder angle of motor 1 ($\theta_1$) and motor 2 ($\theta_2$) in degrees versus time, (b) Position of the colonoscope tip in Cartesian space, and (c) Absolute displacement from the initial position.}
    \label{fig:Experiments}
\end{figure}

\begin{figure}[h!]
   \centering
    \includegraphics[width=1\columnwidth]{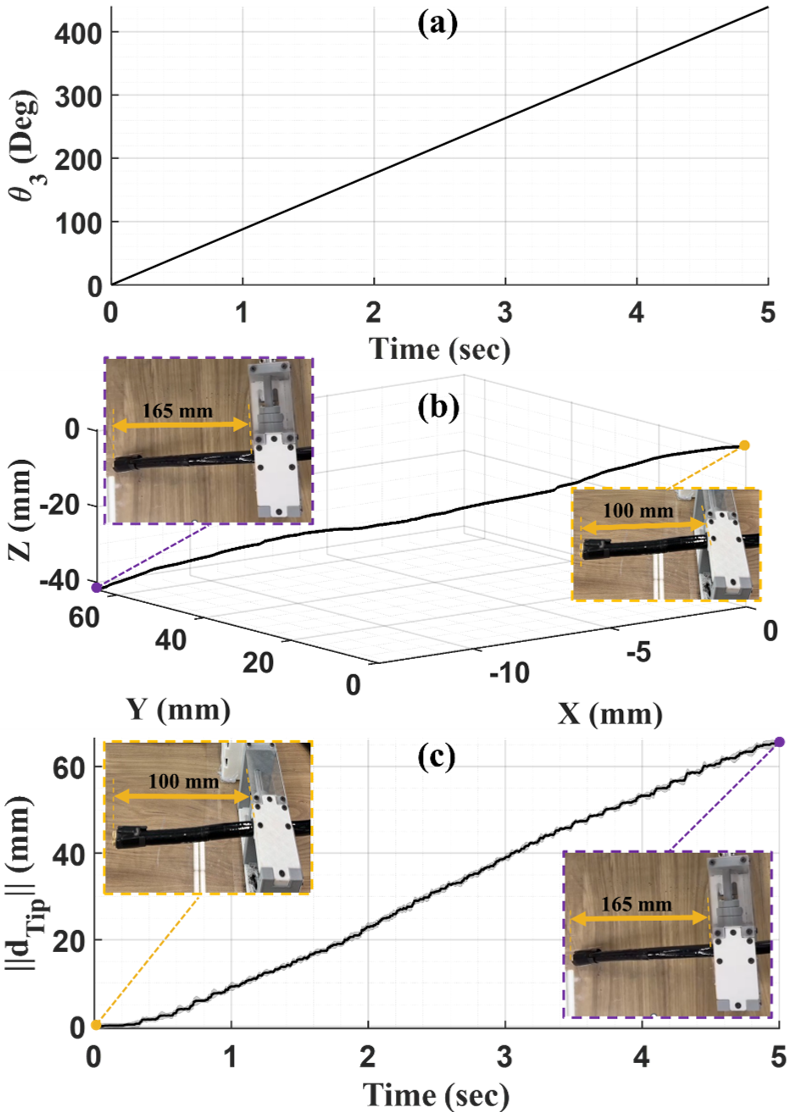}
    \vspace*{-7mm}
    \caption{ Movements of the colonoscope tip measured by a magnetic tracker in insertion using the feeder mechanism. The graphs illustrate: (a) Encoder angle of motor 3 ($\theta_3$) in degrees over time, (b) Colonoscope tip position in Cartesian space, and (c) Absolute displacement from the initial position.}
    \label{fig:Experiments_insertion}
\end{figure}

To evaluate the performance of our proposed concept, we conducted four distinct tests with the colonoscope positioned horizontally and gripped by the feeder mechanism. In each test, we assessed the continuous transmission of movement from motors to the end effector by each collet in CH Gripping mechanism. This was achieved by sending a squared command with a fixed 5-degree increase, while keeping other degrees of freedom constant. For instance, in the first test (red), motor \textit{one} initiated rotation in 5-degree increments to bend the end effector to the right, while the other motors were locked, as shown in Fig.\ref{fig:Experiments}a. To assess the repeatability of the experiments, we conducted each test three times. Then, the mean and standard deviation of tip position in the cartesian space and tip displacement relative to initial straight configuration are calculated. The dark curve and shaded region shown in Fig.\ref{fig:Experiments}b-c represent the mean and standard deviation, respectively. Of note, for assessing repeatability, utilizing a joystick and user for command input lacks precision in repetition. Hence, we generate a command within our C++ code to ensure consistent and accurate testing. Also, magnetic tracker data is recorded at a sampling rate of $40 Hz$ to track the tip of the colonoscope, while encoder data is captured at $20 Hz$ to measure the actual motor position in response to fixed 5-degree increment commands.
Similarly, to evaluate the performance of the Feeder mechanism, we implemented an automated input signal ranging from $\theta_3 = 0^{\circ} $ to $\theta_3 = 440^{\circ}$ degrees to drive the feeder motor, while collecting real-time data from the magnetic tracker. The corresponding results are presented in Fig.\ref{fig:Experiments_insertion}. Of note, during this experiment, the colonoscope tip was left unobstructed to ensure unrestricted forward motion and accurately isolate the feeder mechanism’s performance.

\subsection{\textbf{Case Study}}
\indent To evaluate the practicality of our design in a real scenario, we devised an experiment focused on maneuvering and detecting markers inside a tube, which is shown in Fig.\ref{fig:Experimental Setup}. The objective is to navigate to a final destination and then return to the initial position, demonstrating the feasibility and applicability of our concept.
To conduct the case study, the colonoscope was initially inserted into the tube using the feeder mechanism, activated by pressing the $B_L$ button on the Xbox controller. Once reaching the desired location, the orientation of the colonoscope was precisely adjusted using the bending mechanism, controlled by pressing the $J_L$ axis on the Xbox controller, to detect a marker located at the turning section of the tube. Once the marker was detected, a picture was taken by pressing the $A$ button on the Xbox controller.
After completing the entire screening procedure, the colonoscope was safely retracted using the feeder mechanism and pressing the $B_R$ button on the Xbox controller, with additional guidance provided either manually via the $J_L$ axis or automatically based on the recorded insertion history. This task is demonstrated in Fig.\ref{fig:Trajectory} as performed by a user.
\section{Discussion}
In Fig.\ref{fig:workspace}, despite identical horizontal adaptor displacements ($\Delta$)  applied across all five collet and adaptor geometries (i.e., $S_1$ to $S_5$), distinct deformation behaviors are observed. This clearly highlights the significant influence of the geometrical parameters— such as chuck size and adaptor diameter— on the collet’s performance. For instance, as shown in Fig.\ref{fig:workspace}a–b, increasing the chuck size from $3 mm$ in $S_1$ to $6 mm$ in $S_3$ results in a slight increase in tip deformation. While this increase in tip deflection for a given horizontal input may appear minor, it reflects a noticeable reduction in collet stiffness, thereby lowering the tightening force required to generate the same horizontal movement. From the functionality in the clinic and user perspective, it means that clinicians can open and close the collet with less effort. Additionally, a larger chuck size creates wider gaps between adjacent saw teeth of the collet, allowing for a greater maximum tip displacement. For instance, in sample $S_1$ with a $c = 3 mm$ chuck size, the tip displacement is limited to approximately $2.5 mm$ due to mechanical interference between the saw teeth. In contrast, sample $S_3$, featuring a $c = 6 mm$ chuck size, achieves a tip displacement exceeding $4 mm$.

On the other hand, as shown in Fig.\ref{fig:workspace}c–d, increasing the adaptor diameter from $d = 30 mm$ in $S_4$ to $d = 40 mm$ in $S_6$ leads to a noticeable decrease in tip deformation. For instance, under identical horizontal displacements, the tip deflection nearly doubles when the adaptor diameter is reduced from $40 mm$ to $30 mm$. This behavior is attributed to the increased mechanical advantage achieved with a smaller adaptor diameter, which amplifies the deformation response of the collet. Of note, larger adaptor diameters reduce the force required to achieve a given horizontal displacement due to increased leverage. Therefore, a trade-off must be considered from the functionality perspective of this design: using a smaller adaptor enables quicker assembly with less rotational movement but requires a higher tightening force, while a larger adaptor demands more rotation to achieve the same tip deformation but allows for a reduced tightening effort. Considering this balance, we chose an intermediate adaptor diameter of $d = 34 mm$ —corresponding to case $S_3$— for the proposed design, ensuring a secure grip on the colonoscope’s knob while maintaining ease of use. \\
\indent As shown in Fig.\ref{fig:workspace}, our proposed analytical model accurately predicts the tip displacement of the collet during horizontal movements of the adaptor across all five cases (i.e., $S1 \thicksim S5$). The quantitative comparison is summarized in Table \ref{Samples}, which presents both the maximum error and the mean absolute error for each sample. The worst-case scenario occurs in case $S_4$, where the maximum error and mean absolute error are $ME_4 = 0.72 mm$ and $MAE_4 = 0.44 mm$ , respectively. The discrepancy between the analytical model and FEA results increases when the tip displacement exceeds $2.5 mm$ ($\delta_{\text{Tip}} > 2.5 mm$). This divergence can be attributed to several factors: (i) the analytical model assumes pure bending and neglects shear stresses; (ii) minor geometric mismatches exist between the idealized model and the actual collet structure; and (iii) the analytical approach employs small-deformation theory, whereas the FEA incorporates large deformation effects.

According to the design and simulation analysis —based on our analytical model and verified by ABAQUS— the proposed collet design is capable of securely gripping colonoscope knobs with diameters ranging from $48 mm$ (representing the fully closed state) to $53 mm$ (the collet’s original, undeformed size). This gripping range corresponds to a 67\% increase over the standard clearance of $1.5 mm$, thereby meeting the requirements for quick and efficient assembly. As a result, the collet can accommodate a broad range of clearances, that exists in different commercial colonoscopes, significantly simplifying the disassembly process for technicians during sterilization after each clinical procedure.

Secondly, as shown in Fig.\ref{fig:Experiments}, continuous tip movement was observed in response to continuous input commands during the experimental procedure. The reported mean and standard deviation across three trials indicate consistency, with minor variations primarily attributed to vibrations resulting from the colonoscope’s inherent flexibility during dynamic motion. Distinct bending patterns were observed in both the left–right and up–down directions, as shown in Fig.~\ref{fig:Experiments}b-c. Although the left and right movements appeared symmetrical, Test 4 (green) exhibited a noticeable deviation from Test 3 (black). This deviation can be explained by gravitational effects— opposing upward movements and aiding downward movements. Of note, the distinct patterns observed between left/right and up/down movements are attributed to the different gear ratio in the outer collet (responsible for left/right motion).
\begin{figure}[t]
   \centering
    \includegraphics[width=1\columnwidth]{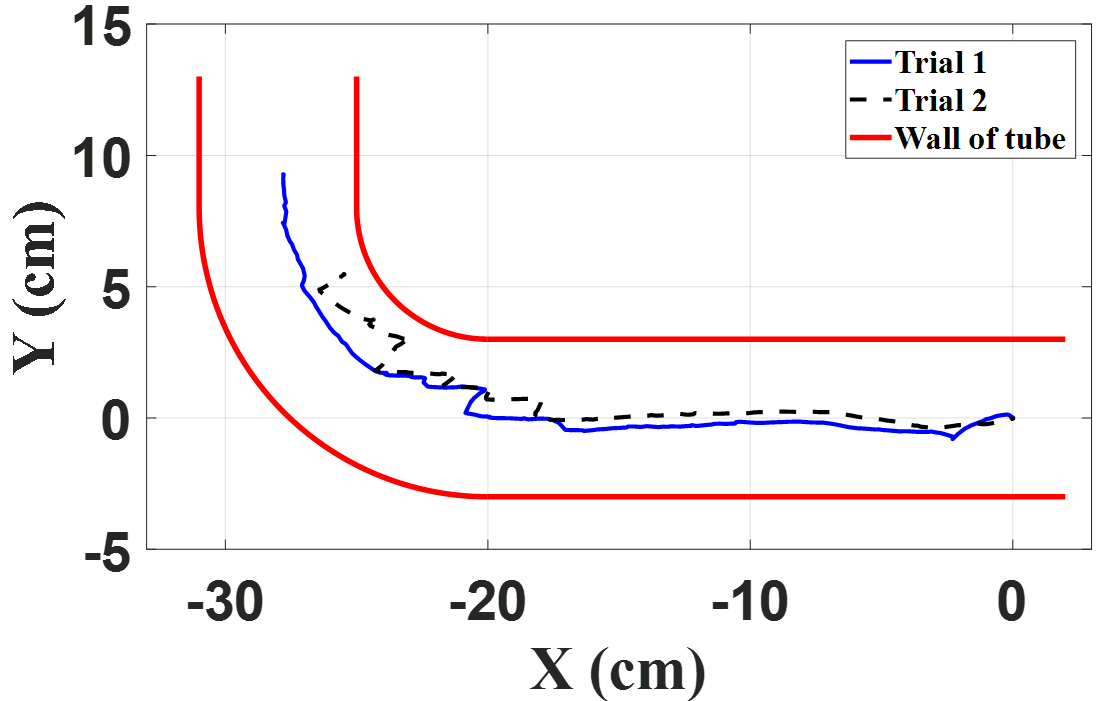}
    \vspace*{-7mm}
    \caption{Trajectory of the colonoscope tip during marker detection.}
    \label{fig:Trajectory}
\end{figure}

As shown in Fig.\ref{fig:Experiments}c, the tip displacement is initially small near the straight configuration and increases progressively as the tip moves further away, reflecting the expected nonlinear response. This graph highlights the inherent nonlinear dynamic behavior of the colonoscope tip, emphasizing the need for extensive surgeon training when using conventional manual systems. In contrast, our proposed setup provides the flexibility to achieve any desired tip behavior through precise background control, enabling faster and more intuitive training. Surgeons no longer need to continuously adjust knob rotation to account for the system’s nonlinearity, allowing them to focus on cancer diagnosis rather than managing the mechanical complexity of the colonoscope. 
As shown in Fig.\ref{fig:Experiments_insertion}, the feeder mechanism is able to advance the colonoscope smoothly and effectively in the absence of external obstructions. This confirms the mechanism’s capability to deliver consistent and continuous motion under ideal conditions, thereby validating the baseline performance of the proposed design. 

Finally, as depicted in Fig.\ref{fig:Trajectory}, users can simply press the $B_L$ button to activate the feeder mechanism and guide the colonoscope tip using the $J_L$ axis on an Xbox controller. This intuitive control scheme significantly can reduce the learning curve typically associated with endoscopic navigation. In contrast, traditional manual control of a colonoscope requires precise two-handed coordination to manipulate and lock/unlock mechanical knobs across multiple degrees of freedom. This complex process demands significant training and often leads to user fatigue, inefficient maneuvers, and potential damage to the colonoscope or surrounding tissue \cite{Fisher2011ComplicationsOC,
shine2020quality}. By simplifying these complex actions into an intuitive joystick-based interface, our system can enhance usability, lower cognitive load, and facilitate safer, more effective navigation for both novice and experienced users. 
In real clinical applications, the system’s automatic mode can further assist surgeons by automatically bending the tip in the required direction, locking it in place, and dynamically adjusting to follow the movement of polyps. This ensures uninterrupted visualization during biopsy and allows the surgeon to focus fully on polyp removal without needing to manage the colonoscope’s orientation manually. 

\section{Concluding Remarks}
The proposed mechanism demonstrates a notable advantage through its seamless integration with existing commercialized colonoscopic and endoscopic systems, ensuring both affordability and practical deployment. This integration eliminates the need for custom fabrication of new scopes, thereby not only reducing development costs but also simplifying the FDA approval process. Additionally, it can simplify procedures and assists surgeons in the early detection of cancer and polyps in the GI tract by potentially reducing workload, fatigue, and distractions— thanks to an intuitive, modular, and easily installable mechatronics system and intuitive user interface.
Furthermore, our analytical model shows strong agreement with FEA results, with errors below 10\% in the operational region. This level of accuracy eliminates the need for time-consuming FEA simulations when designing collets for other endoscopic tools.
Experimental evaluations confirm effective torque transmission from the motor to the colonoscope knob, enabling precise tip control. The insertion mechanism uses friction-based feeding, ensuring sufficient insertion speed while preventing excessive force application to the phantom. 

In future work, we plan to conduct extensive user studies to quantitatively evaluate the learning curve of our system compared to the traditional systems while performing a colonoscopy procedure. Additionally, we aim to implement remote assistance functionality, enabling operation of the colonoscope from different locations. We also intend to integrate complementary machine learning and artificial intelligence for an enhanced colorectal cancer screening and diagnosis.

\ifCLASSOPTIONcaptionsoff
  \newpage
\fi

\addtolength{\textheight}{-1cm}   

\bibliographystyle{IEEEtran}
\bibliography{main}

\begin{thebibliography}{10}
\providecommand{\url}[1]{#1}
\csname url@samestyle\endcsname
\providecommand{\newblock}{\relax}
\providecommand{\bibinfo}[2]{#2}
\providecommand{\BIBentrySTDinterwordspacing}{\spaceskip=0pt\relax}
\providecommand{\BIBentryALTinterwordstretchfactor}{4}
\providecommand{\BIBentryALTinterwordspacing}{\spaceskip=\fontdimen2\font plus
\BIBentryALTinterwordstretchfactor\fontdimen3\font minus \fontdimen4\font\relax}
\providecommand{\BIBforeignlanguage}[2]{{%
\expandafter\ifx\csname l@#1\endcsname\relax
\typeout{** WARNING: IEEEtran.bst: No hyphenation pattern has been}%
\typeout{** loaded for the language `#1'. Using the pattern for}%
\typeout{** the default language instead.}%
\else
\language=\csname l@#1\endcsname
\fi
#2}}
\providecommand{\BIBdecl}{\relax}
\BIBdecl

\bibitem{Williams2009}
J.~Waye, D.~Rex, and C.~Williams, ``Colonoscopy: Principles and practice, second edition,'' \emph{Colonoscopy: Principles and Practice, Second Edition}, pp. 1--816, 09 2009.

\bibitem{sung2021global}
H.~Sung, J.~Ferlay, R.~L. Siegel, M.~Laversanne, I.~Soerjomataram, A.~Jemal, and F.~Bray, ``Global cancer statistics 2020: Globocan estimates of incidence and mortality worldwide for 36 cancers in 185 countries,'' \emph{CA: a cancer journal for clinicians}, vol.~71, no.~3, pp. 209--249, 2021.

\bibitem{zhao2019magnitude}
S.~Zhao, S.~Wang, P.~Pan, T.~Xia, X.~Chang, X.~Yang, L.~Guo, Q.~Meng, F.~Yang, W.~Qian \emph{et~al.}, ``Magnitude, risk factors, and factors associated with adenoma miss rate of tandem colonoscopy: a systematic review and meta-analysis,'' \emph{Gastroenterology}, vol. 156, no.~6, pp. 1661--1674, 2019.

\bibitem{azer2019challenges}
S.~A. Azer, ``Challenges facing the detection of colonic polyps: What can deep learning do?'' \emph{Medicina}, vol.~55, no.~8, p. 473, 2019.

\bibitem{Alian2022CurrentED}
A.~Alian, E.~Zari, Z.~Wang, E.~Franco, J.~Avery, M.~S. Runciman, B.~Lo, F.~R. y~Baena, and G.~P. Mylonas, ``Current engineering developments for robotic systems in flexible endoscopy,'' \emph{Techniques and Innovations in Gastrointestinal Endoscopy}, 2022.

\bibitem{Gerald2022ASR}
A.~Gerald, R.~Batliwala, J.~Ye, P.~Hsu, H.~Aihara, and S.~Russo, ``A soft robotic haptic feedback glove for colonoscopy procedures,'' \emph{2022 IEEE/RSJ International Conference on Intelligent Robots and Systems (IROS)}, pp. 583--590, 2022.

\bibitem{Kim2023ASA}
H.~S. Kim, O.~C. Kara, and F.~Alambeigi, ``A soft and inflatable vision-based tactile sensor for inspection of constrained and confined spaces,'' \emph{IEEE Sensors Journal}, vol.~23, pp. 29\,605--29\,618, 2023.

\bibitem{Kume2015DevelopmentOA}
K.~Kume, N.~Sakai, and T.~Goto, ``Development of a novel endoscopic manipulation system: the endoscopic operation robot ver.3,'' \emph{Endoscopy}, vol.~47, pp. 815 -- 819, 2015.

\bibitem{Takamatsu2023RoboticEW}
T.~Takamatsu, Y.~Endo, R.~Fukushima, T.~Yasue, K.~Shinmura, H.~Ikematsu, and H.~Takemura, ``Robotic endoscope with double-balloon and double-bend tube for colonoscopy,'' \emph{Scientific Reports}, vol.~13, 2023.

\bibitem{Reilink2010EndoscopicCC}
R.~Reilink, G.~de~Bruin, M.~Franken, M.~A. Mariani, S.~Misra, and S.~Stramigioli, ``Endoscopic camera control by head movements for thoracic surgery,'' \emph{2010 3rd IEEE RAS \& EMBS International Conference on Biomedical Robotics and Biomechatronics}, pp. 510--515, 2010.

\bibitem{Vrielink2018IntuitiveGO}
T.~J. C.~O. Vrielink, J.~G.-B. Puyal, A.~A. Kogkas, A.~Darzi, and G.~P. Mylonas, ``Intuitive gaze-control of a robotized flexible endoscope,'' \emph{2018 IEEE/RSJ International Conference on Intelligent Robots and Systems (IROS)}, pp. 1776--1782, 2018.

\bibitem{Lee2020easyEndoRE}
D.-H. Lee, B.~Cheon, J.~Kim, and D.-S. Kwon, ``easyendo robotic endoscopy system: Development and usability test in a randomized controlled trial with novices and physicians,'' \emph{The International Journal of Medical Robotics and Computer Assisted Surgery}, vol.~17, 2020.

\bibitem{Basha2023AGS}
S.~Basha, M.~Khorasani, N.~Abdurahiman, J.~Padhan, V.~M. Baez, A.~A. Al-Ansari, P.~Tsiamyrtzis, A.~T. Becker, and N.~V. Navkar, ``A generic scope actuation system for flexible endoscopes,'' \emph{Surgical Endoscopy}, vol.~38, pp. 1096 -- 1105, 2023.

\bibitem{Iwasa2018ANR}
T.~Iwasa, R.~Nakadate, S.~Onogi, Y.~Okamoto, J.~Arata, S.~Oguri, H.~Ogino, E.~Ihara, K.~Ohuchida, T.~Akahoshi, T.~Ikeda, Y.~Ogawa, and M.~Hashizume, ``A new robotic-assisted flexible endoscope with single-hand control: endoscopic submucosal dissection in the ex vivo porcine stomach,'' \emph{Surgical Endoscopy}, vol.~32, pp. 3386--3392, 2018.

\bibitem{Chapra2001NumericalMF}
S.~C. Chapra, ``Numerical methods for engineers: with software and programming applications / steven c. chapra, raymond p. canale,'' 2001.

\bibitem{HEARN1997254}
E.~HEARN, ``Chapter 11 - strain energy,'' in \emph{Mechanics of Materials 1 (Third Edition)}, third edition~ed., E.~HEARN, Ed.\hskip 1em plus 0.5em minus 0.4em\relax Oxford: Butterworth-Heinemann, 1997, pp. 254--296.

\bibitem{Fisher2011ComplicationsOC}
D.~A. Fisher, J.~T. Maple, T.~Ben-Menachem, B.~D. Cash, G.~A. Decker, D.~Early, J.~A. Evans, R.~D. Fanelli, N.~Fukami, J.~H. Hwang, R.~Jain, T.~L. Jue, K.~M. Khan, P.~M. Malpas, R.~N. Sharaf, A.~K. Shergill, and J.~A. Dominitz, ``Complications of colonoscopy.'' \emph{Gastrointestinal endoscopy}, vol. 74 4, pp. 745--52, 2011.

\bibitem{shine2020quality}
R.~Shine, A.~Bui, and A.~Burgess, ``Quality indicators in colonoscopy: an evolving paradigm,'' \emph{ANZ journal of surgery}, vol.~90, no.~3, pp. 215--221, 2020.

\end{thebibliography}








\end{document}